\documentclass[sigconf]{acmart}
\renewcommand\footnotetextcopyrightpermission[1]{} % removes footnote with conference information in first column
\usepackage{multirow} 
\usepackage{booktabs}
\usepackage{threeparttable}
\usepackage{footnote}
\AtBeginDocument{%
  \providecommand\BibTeX{{%
    \normalfont B\kern-0.5em{\scshape i\kern-0.25em b}\kern-0.8em\TeX}}}

\setcopyright{acmcopyright}
\copyrightyear{2018}
\acmYear{2018}
\acmDOI{XXXXXXX.XXXXXXX}

\acmConference[Conference acronym 'XX]{Make sure to enter the correct
  conference title from your rights confirmation emai}{June 03--05,
  2018}{Woodstock, NY}
\acmPrice{15.00}
\acmISBN{978-1-4503-XXXX-X/18/06}

\begin{document}

%%
%% The "title" command has an optional parameter,
%% allowing the author to define a "short title" to be used in page headers.
\title{Self-recoverable Adversarial Examples: A New Effective Protection Mechanism in Social Networks}

\author{Jiawei Zhang}
\affiliation{%
	\institution{Nanjing University of Information Science and Technology}
	\city{Nanjing}
	\country{China}}
\email{zjwei\_2020@nuist.edu.cn}

\author{Jinwei Wang}
\affiliation{%
	\institution{Nanjing University of Information Science and Technology}
	\city{Nanjing}
	\country{China}}
\email{wjwei\_2004@163.com}

\author{Hao Wang}
\affiliation{%
	\institution{Nanjing University of Science and Technology}
	\city{Nanjing}
	\country{China}}
\email{sa875923372@163.com}

\author{Xiangyang Luo}
\affiliation{%
	\institution{State Key Laboratory of Mathematical Engineering and Advanced Computing}
	\city{Zhenzhou}
	\country{China}}
\email{xiangyangluo@126.com}

%%
%% By default, the full list of authors will be used in the page
%% headers. Often, this list is too long, and will overlap
%% other information printed in the page headers. This command allows
%% the author to define a more concise list
%% of authors' names for this purpose.
\renewcommand{\shortauthors}{Jiawei Zhang and Jinwei Wang, et al.}

%%
%% The abstract is a short summary of the work to be presented in the
%% article.
\begin{abstract}
Malicious intelligent algorithms greatly threaten the security of social users’ privacy by detecting and analyzing the uploaded photos to social network platforms. The destruction to DNNs brought by the adversarial attack sparks the potential that adversarial examples serve as a new protection mechanism for privacy security in social networks. However, the existing adversarial example does not have recoverability for serving as an effective protection mechanism. To address this issue, we propose a recoverable generative adversarial network to generate $self$-$recoverable$ $adversarial$ $examples$. By modeling the adversarial attack and recovery as a united task, our method can minimize the error of the recovered examples while maximizing the attack ability, resulting in better recoverability of adversarial examples. To further boost the recoverability of these examples, we exploit a dimension reducer to optimize the distribution of adversarial perturbation. The experimental results prove that the adversarial examples generated by the proposed method present superior recoverability, attack ability, and robustness on different datasets and network architectures, which ensure its effectiveness as a protection mechanism in social networks.
\end{abstract}

%%
%% The code below is generated by the tool at http://dl.acm.org/ccs.cfm.
%% Please copy and paste the code instead of the example below.
%%
\begin{CCSXML}
	<ccs2012>
	<concept>
	<concept_id>10002978.10003029.10003032</concept_id>
	<concept_desc>Security and privacy~Social aspects of security and privacy</concept_desc>
	<concept_significance>500</concept_significance>
	</concept>
	<concept>
	<concept_id>10002978.10003029.10011150</concept_id>
	<concept_desc>Security and privacy~Privacy protections</concept_desc>
	<concept_significance>500</concept_significance>
	</concept>
	<concept>
	<concept_id>10010147.10010178.10010224.10010245.10010251</concept_id>
	<concept_desc>Computing methodologies~Object recognition</concept_desc>
	<concept_significance>500</concept_significance>
	</concept>
	</ccs2012>
\end{CCSXML}

\ccsdesc[500]{Security and privacy~Social aspects of security and privacy}
\ccsdesc[500]{Security and privacy~Privacy protections}
\ccsdesc[500]{Computing methodologies~Object recognition}

%%
%% Keywords. The author(s) should pick words that accurately describe
%% the work being presented. Separate the keywords with commas.
\keywords{Social networks, deep learning, adversarial attack and recover, privacy protection}

%% A "teaser" image appears between the author and affiliation
%% information and the body of the document, and typically spans the
%% page.
%%
%% This command processes the author and affiliation and title
%% information and builds the first part of the formatted document.
\maketitle

\section{Introduction}
\label{Introduction}
Deep neural networks (DNNs) have achieved excellent performance in many tasks, such as image processing and semantic recognition. However, recent works show that DNNs are vulnerable to adversarial examples. The adversarial examples can be generated by adding some special and imperceptible noise to the normal examples, and it can make the target DNNs output the wrong predictions.
\begin{figure}[ht]
	\centering
	\includegraphics[scale=0.38]{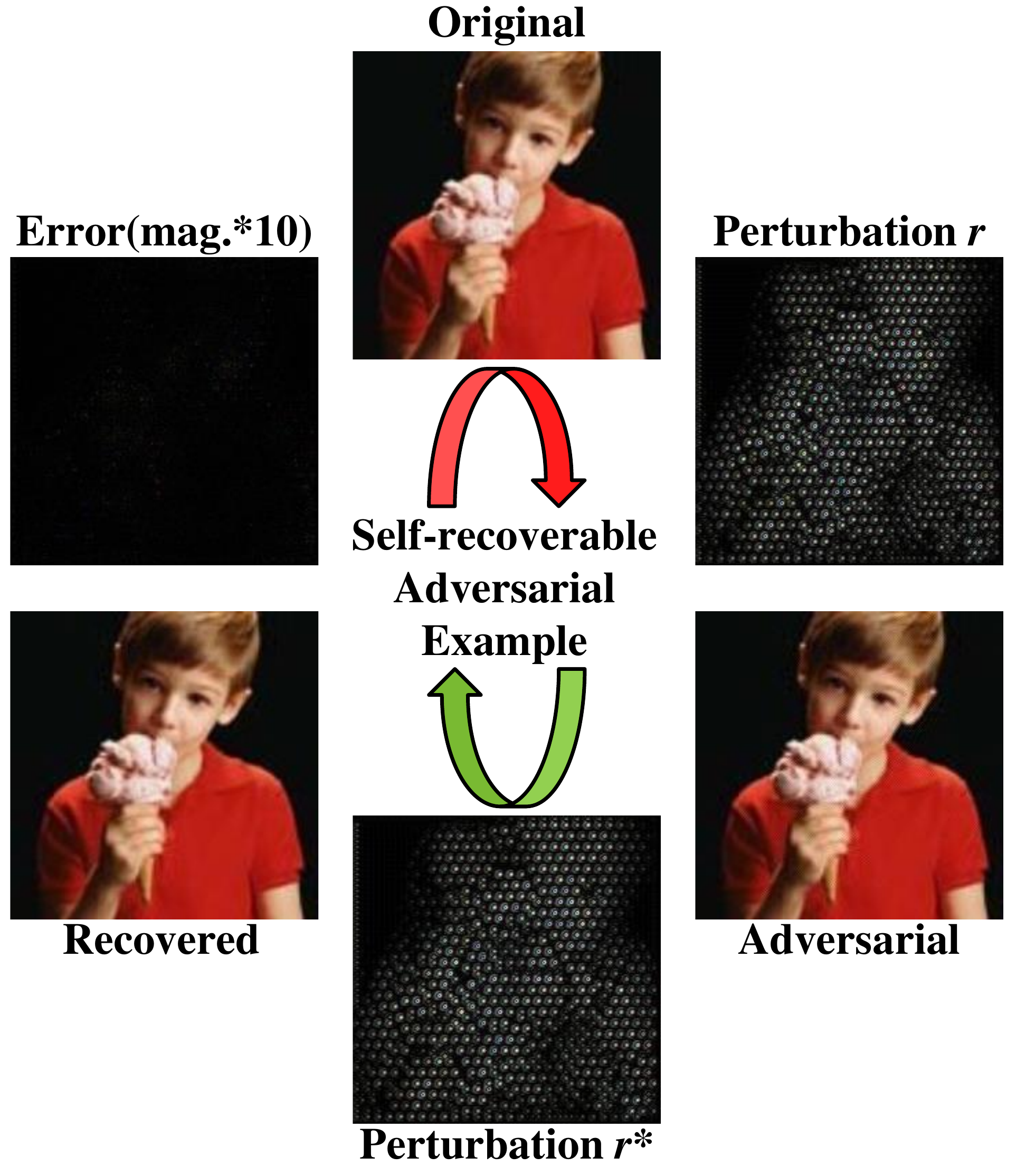}
	\caption{By modeling adversarial attack and recovery as a united task, the $self$-$recoverable$ $adversarial$ $example$ (SRAE) can minimize the error of the recovered example while maximizing the attack ability. The SRAE can serve as an effective protection mechanism against malicious DNNs while remaining harmless to ourselves.}
	\label{demo1}
\end{figure}
\begin{figure*}[ht]
	\centering
	\includegraphics[scale=0.66]{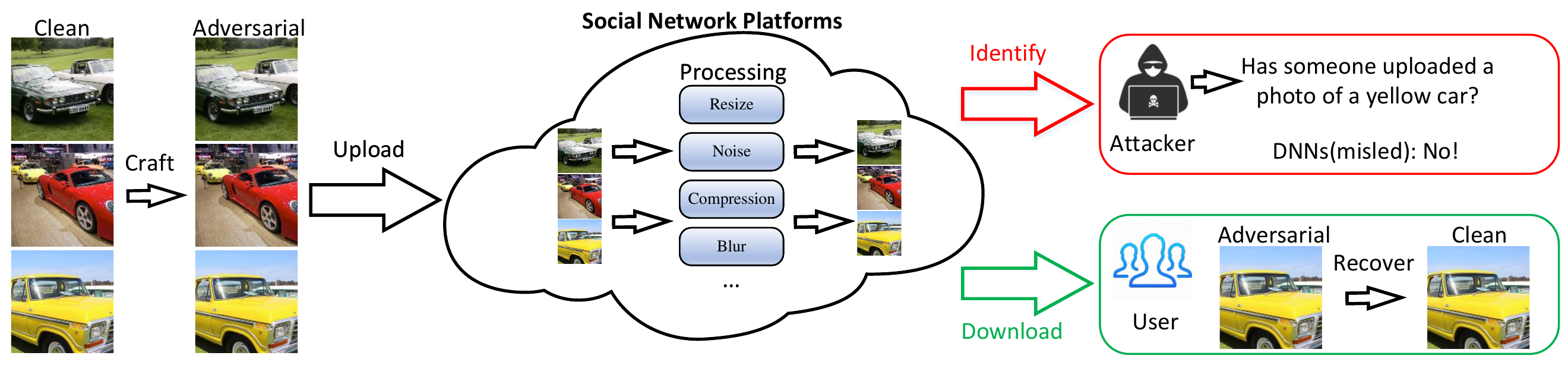}
	\caption{The SRAE can serve as a new protection mechanism against attackers on social network platforms. By crafting the uploaded images to adversarial examples, the malicious intelligent algorithms (e.g., DNNs) fail to identify the image content. This protection mechanism can prevent users' social relations, property, and identity from being collected and analyzed.}	
	\label{sense}
\end{figure*}

The generation manner of adversarial examples can be categorized into white-box and black-box. With the knowledge of the structure and parameters of the targeted DNNs, the adversarial examples \cite{lbfgs,cw,fgsm,ifgsm,madry2017towards,dong2018boosting,rony2019decoupling,xiao2018generating,Liu_2019_CVPR} can be generated in a white-box manner, including the optimization-based method L-BFGS \cite{lbfgs}, the gradient-based methods FGSM \cite{fgsm} and various iterative variants \cite{ifgsm,dong2018boosting,xie2019improving,cheng2019improving}. Besides, due to the transferability \cite{tang2021robustart}, the adversarial examples can perform attacks in a black-box manner. For example, the adversarial examples generated according to model $A$ can also mislead model $B$, even if the structure and parameters of $A$ are different from $B$. Although the existing adversarial defense methods \cite{wu2019defending,jin2019ape,liao2018defense,zhou2021towards,liu2019detection,wang2021smsnet,liang2018detecting,grosse2017statistical,gong2017adversarial,wu2020adversarial} can improve DNNs' robustness, some new adversarial examples can always break these defense methods. As a result, the adversarial examples are practical in real-world conditions and pose a great threat to the reliability of DNNs. 

Although adversarial examples bring destruction and threat to DNNs, we try to exploit these negative impacts to fulfill the positive potential of these examples serving as a new protection mechanism for privacy security in social networks. Specifically, nowadays, users upload numerous photos to social network platforms for sharing their daily lives. These photos contain personal private information, including users’ social relationships, properties, identities. This information can be easily detected and collected by malicious intelligent algorithms (e.g., DNNs), which greatly threatens the security of social users’ privacy \cite{tonge2020image, tonge2018identifying, zhong2017group}. Therefore, we aim to protect image privacy based on the theory of adversarial examples. Compared with other protection methods, crafting an image as an adversarial example are imperceptible and can effectively prevent the prevailing DNNs from detecting, classification, and further analyzing the image content.

However, the existing adversarial attack lacks the study on recoverability and reversibility, which makes them unable to serve as an effective protection mechanism. Therefore, as shown in Figure \ref{demo1}, we consider crafting a $self$-$recoverable$ $adversarial$ $examples$ (SRAE), which owns high attack ability under various cases (e.g., disturbance, adversarial defense) and can only be recovered near losslessly by ourselves. Based on this, SRAE can serve as a new protection mechanism (shown in Figure \ref{sense}) against attackers and avoid the data being identified, collected, and analyzed by malicious DNNs while remaining harmless to users. Moreover, the transferability of adversarial examples can give a great generalization ability to this protection mechanism, making it still effective in the unaware adversarial environment. 

In this paper, we propose a $recoverable$ $generative$ $adversarial$ $network$ (RGAN) to generate the proposed SRAE in an end-to-end way. Beyond the existing white-box attacks, which constantly need the gradient information of target DNNs, the proposed RGAN is free from the structure and parameters. More importantly, instead of treating the adversarial attack and recovery as two separate and independent tasks, we attempt to model the attack and recovery as a united task by the proposed framework. For the purpose of learning the distribution of adversarial perturbation better, the recovery part is specially designed and trained jointly and dynamically with the generation part in a pipeline. The experimental results show that the SRAE has better recoverability on different datasets and network architectures. SRAE minimizes the error of the recovered example while maximizing the attack ability, and outperforms the combinations of existing adversarial attack and defense methods.

To further boost the recoverability of SRAE, we study the relationship between the recovered error and the distribution of adversarial perturbation. Given network architecture with certain simulation capabilities, we observe that adversarial perturbations with lower intensity or simpler structures are easier to be recovered. Therefore, we design a dimension reducer to optimize the distribution of adversarial perturbation. We show that the SRAE optimized with the proposed dimension reducer can be recovered to original examples with negligible error, further satisfying the harmless requirement of recovered examples. To summarize, the main contributions of this paper are:

\begin{itemize}
	\item Our proposed RGAN is the first attempt to model adversarial attack and recovery, a pair of mutually-inverse challenges, as a united task. Powered by joint dynamic training, the recoverable adversarial examples maximize the attack ability and can be recovered near losslessly by our RGAN.
	\item We study the effects of adversarial perturbation distribution on recoverability, demonstrating that perturbation with lower intensity or simpler structures is easier to recover. Therefore, we design a dimension reducer to optimize the perturbation distribution, further boosting the recoverability.
	\item The experimental results show that the proposed method presents superior recoverability than the combinations of state-of-the-art attack and defense methods on different datasets and network architectures, which can serve as a new effective protection mechanism for privacy security in social networks.
\end{itemize}

\begin{figure*}[ht]
	\centering
	\includegraphics[scale=0.35]{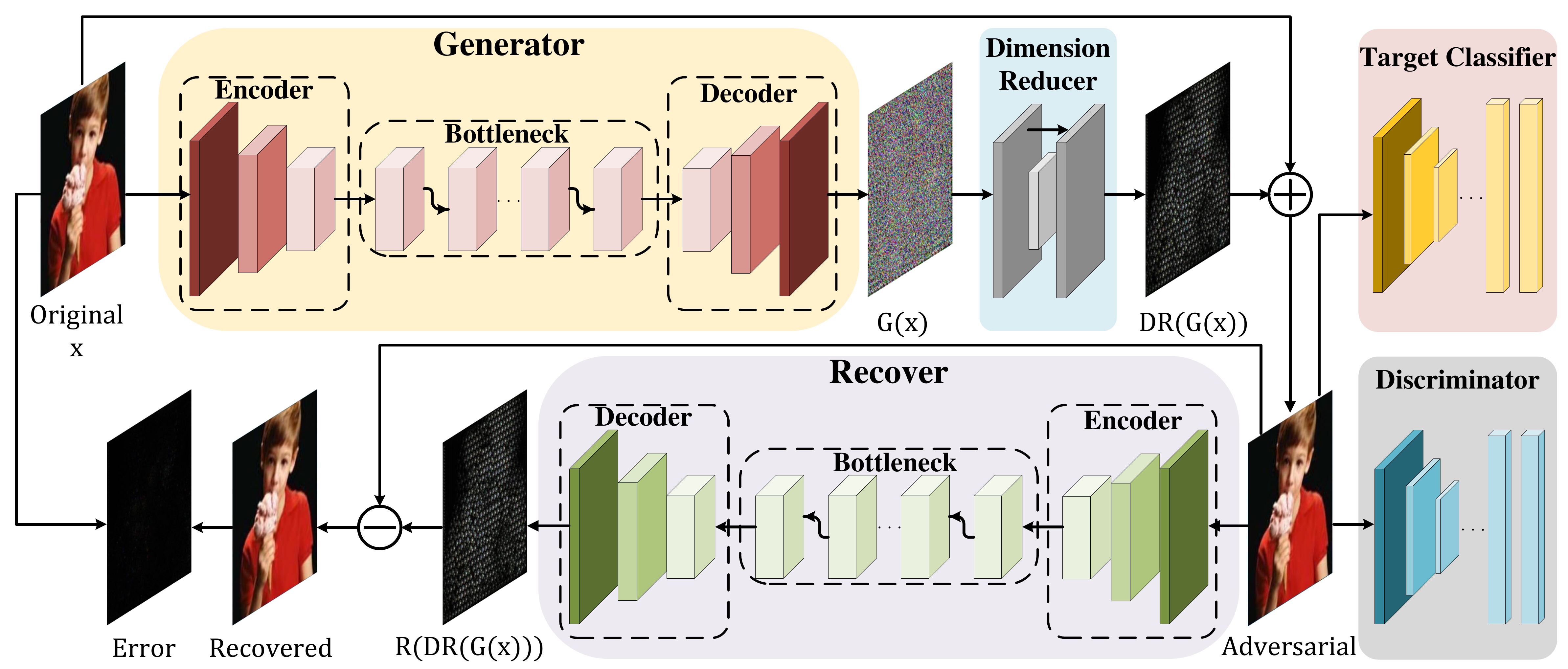}
	\caption{The overall framework of our $recoverable$ $generative$ $adversarial$ $network$ (RGAN). Our main idea is to model the adversarial attack and recovery as a united task. By optimizing the perturbation distribution with the dimension reducer, the SRAE can achieve high attack ability and recoverability to form a powerful protection mechanism.}	
	\label{framework}
\end{figure*}

\section{Related Work}
In this section, we briefly describe the notion used in this paper. Furthermore, due to the lack of recoverable adversarial examples, we review some related works about adversarial attack and defense. In Section \ref{Experiment and Analysis}, these attacks and defense will be combined to serve as competitive solutions with our SRAE. 

We denote $x$ as the clean image from the dataset and $y$ as the ground-truth label of $x$. A target deep neural network is represented by model $f(\cdot)$, which can achieve $f(x)=y$. The goal of adversarial attack is to find a perturbation $r$, which meet $f(x+r) \neq y$. Let $x^{adv}$ represent adversarial example, which means $x^{adv}=x+r$.

\subsection{Existing Methods for Adversarial Attack}
Existing methods for generating adversarial examples can be categorized into three groups: optimization-based, gradient-based, and generation-based.

\textbf{Optimization-based methods} \cite{lbfgs,cw} solve the generation of adversarial example as an optimization problem, which combines the magnitude of perturbation $r$ with the attack ability of adversarial example $x^{adv}$ as the optimization goal \cite{lbfgs}. 

\textbf{Gradient-based methods} \cite{fgsm,ifgsm,madry2017towards,dong2018boosting,rony2019decoupling} maximize the loss function by a chosen perturbation step size $\epsilon$ according to the gradient direction \cite{fgsm}. Although the gradient-based methods have certain drawbacks in attack ability, they are much faster than the optimization-based method. 

\textbf{Generation-based method} \cite{xiao2018generating,Liu_2019_CVPR} train another model to generate perturbation for the target model. These methods have similar flexibility as the optimization-based method but cost less time. Moreover, these methods can generate adversarial examples without the parameters and structure of the target model. 

\subsection{Existing Methods for Defenses}
Existing methods for adversarial defense can also be categorized into three main directions: adversarial training, adversarial denoising, and adversarial detection.

\textbf{Adversarial training} \cite{fgsm,tramer2017ensemble,madry2017towards,wu2019defending,wu2020adversarial} augment the training dataset with adversarial examples to train a robust model. 

\textbf{Adversarial denoising} \cite{liao2018defense,jin2019ape,zhou2021towards} exploit a denoiser to pre-process the input, which can erase the aggressive of adversarial examples.

\textbf{Adversarial detection} \cite{liu2019detection,wang2021smsnet,liang2018detecting,grosse2017statistical,gong2017adversarial} aim at determine whether the input is adversarial or not, rather than eliminating the adversarial property.

\section{Proposed Method}

\subsection{Overview}
Our goal is to develop a learnable, end-to-end model for $self$-$recoverable$ $adversarial$ $examples$ (SRAE) that can form a protection mechanism. Due to the combination of adversarial property and high recoverability, SRAE can be aggressive to the attacker while being harmless to ourselves. Note that the recoverability doesn't mean the SRAE are fragile and can be easily destroyed. In opposite, the SRAE are robust to various transformations $T(\cdot)$ existing in social networks and other adversarial defense methods (e.g., JPEG compression, Gaussian noise, denoising filter, APE\cite{jin2019ape}, ARN\cite{zhou2021towards}). In our scenario, SRAE can only be precisely recovered by the proposed $recoverable$ $generative$ $adversarial$ $network$ (RGAN).

As shown in Figure \ref{framework}, the proposed RGAN consists of five parts: a generator $G$, a dimension reducer $DR$, a discriminator $D$, a target classifier $C$, and a recover $R$. Specifically, the generator $G$ takes the original examples $x$ as the input and outputs the perturbations $G(x)$. Then, the perturbations $G(x)$ are optimized with the dimension reducer $DR$. Afterward, the adversarial examples can be obtained by $x+DR(G(x))$. Next, the adversarial examples are sent to the discriminator $D$ and the target classifier $C$ for indistinguishability and aggressiveness optimization. Meanwhile, the adversarial examples are sent to the recover $R$, which aims to recover these examples to the original examples. It is worth noting that the proposed framework is different from the existing denoising methods. The proposed RGAN jointly trains the generator $G$ and the recover $R$, rather than separating the recovery from the attack, resulting in a better recoverability. This allows the generated SRAE to serve as a protection mechanism for privacy security in social networks.

\subsection{Recoverable Generative Adversarial Network (RGAN)}

\textbf{The generator $G$} is the starting point and aims to generate perturbation according to the features of the input $x$. It consists of an encoder, a bottleneck module, and a decoder. The encoder, consisting of three convolution layers, normalization, and $ReLU$ activation function, extracts features from the clean images. Correspondingly, the decoder exploits deconvolution layers, normalization, and activation functions to map the features to perturbation with the same size as the image. To increase the representative capacity of the generator $G$, we add the bottleneck module between the encoder and the decoder, which consists of residual blocks \cite{he2016deep}.

\textbf{The recover $R$} aims to recover the adversarial examples. We study the effect of different structures of the recover $R$ towards the recoverability. Suppose the recover $R$ is more complex and deeper than the generator $G$, whether it can better recover the adversarial examples. 
\begin{figure}[ht]
	\centering
	\includegraphics[scale=0.6]{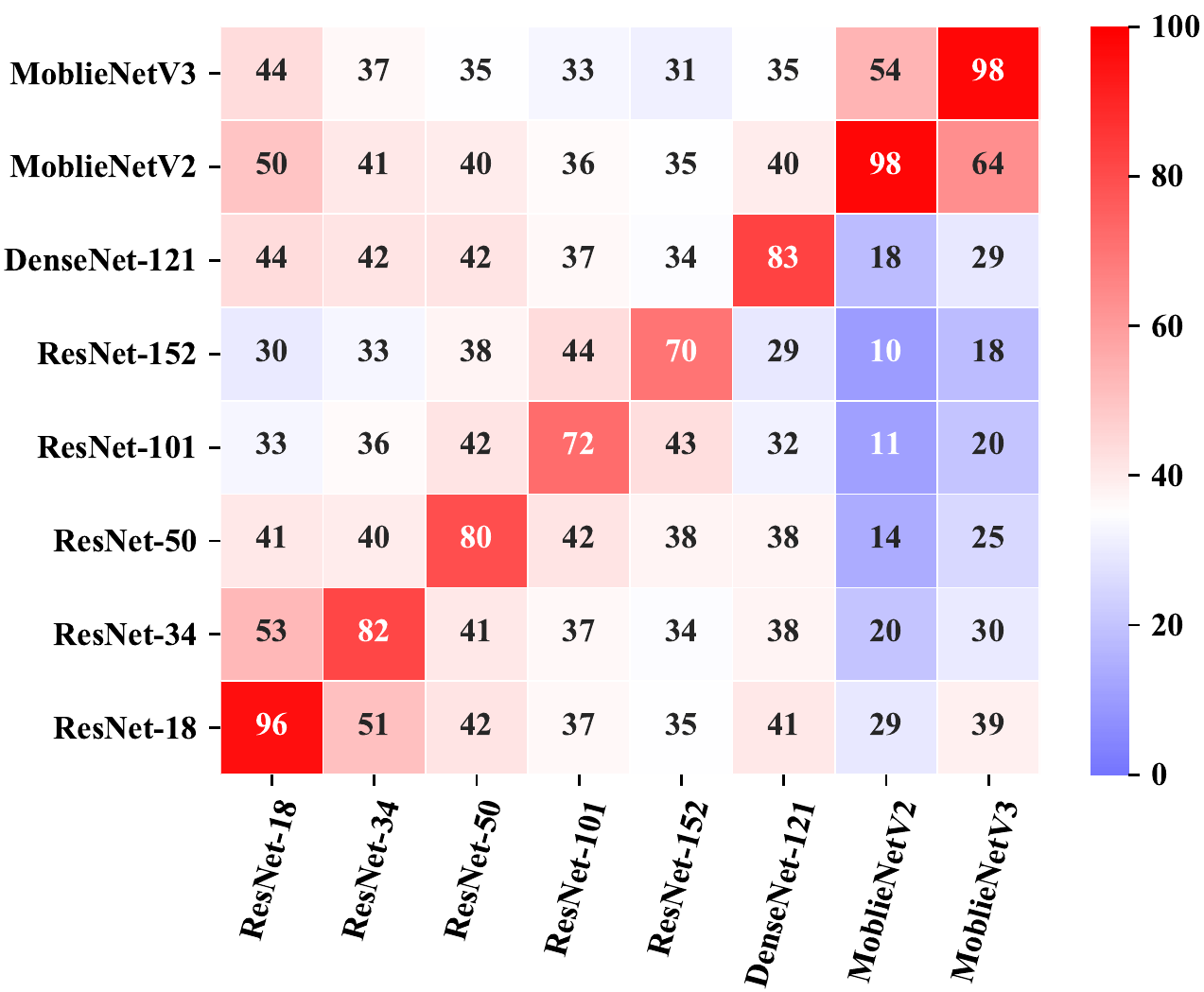}
	\caption{Transferability heatmap of adversarial examples generated by FGSM ($\epsilon$=8/255). Values represent attack success rates ($\%$) of the examples against a target model (column), which is generated according to a source model (row).}
	\label{transfer}
\end{figure}

As shown in Figure \ref{transfer}, diving into the transferability of adversarial examples, we find that the examples generated for a specific network transfer better to its homologous networks. For example, the adversarial examples generated for one ResNet are more likely to be adversarial to another network within the ResNet family. Moreover, within the network family, the adversarial examples generated for networks with similar depth tend to have better transferability. These phenomenons reveal that \textbf{the homologous networks with similar depth tend to have more similar decision boundaries.}

Based on the above observation, we settle the recover $R$ to be a homologous network with the same depth of the generator $G$. With a similar structure and depth, the recover $R$ can better extract the perturbation generated by the generator $G$. We study the recoverability brought by different depths of the generator $G$ and the recover $R$ in Section \ref{network structure}.

\textbf{The dimension reducer $DR$} is the essential part of RGAN, which further boosts the recoverability of SRAE. Specifically, through a down-sample and up-sample operation, we optimize the perturbation distribution of SRAE, resulting in the reduction of intensity and complexity. As briefly described in the introduction part, the dimension reducer $DR$ is motivated by the following two observations:
\begin{itemize}
	\item{\textbf{\textit{The perturbation generated by the generation-based m-ethod is larger and messier than the gradient-based and optimization-based method.}}}	
	\item{\textbf{\textit{The less complex perturbation is easier to be recovered.}}}
\end{itemize}

\begin{figure}[ht]
	\centering
	\includegraphics[scale=0.18]{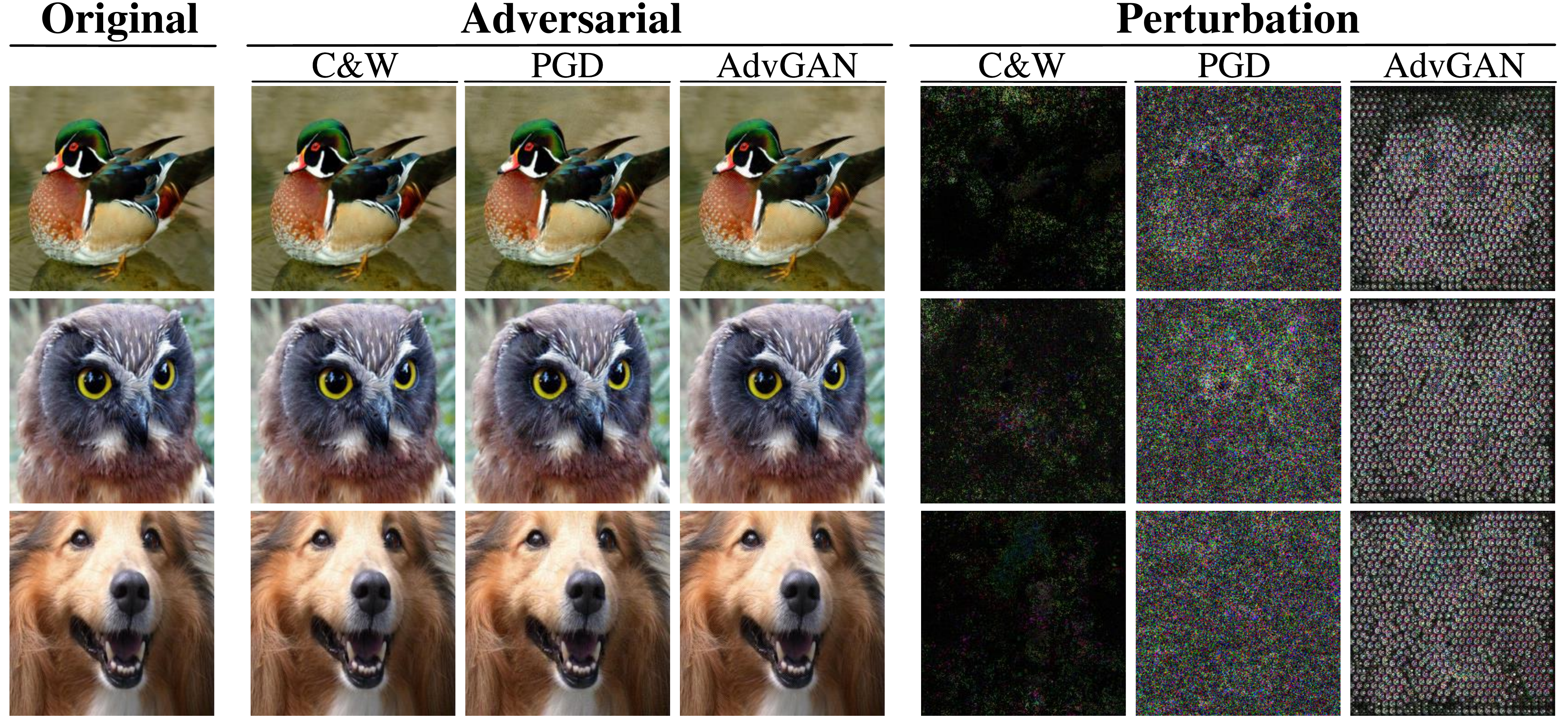}
	\caption{Examples generated by various methods. From the left: original example, adversarial examples, adversarial perturbation. From the left in adversarial and perturbation: the optimization-based C\&W, the gradient-based PGD, the generation-based AdvGAN.}
	\label{noise diff}
\end{figure}

As shown in Figure \ref{noise diff}, \textbf{\textit{the perturbation generated by the gen-\\eration-based method is larger and messier than the gradient-based and optimization-based method.}} This observation pushes us to consider how to reduce the redundancy of perturbation generated by the generation-based method. Simply magnifying the weight of perturbation intensity in the loss function does help to reduce the perturbation intensity. However, it brought certain drawbacks to the attack ability (see appendix \ref{discussion on reducing}). Therefore, we focus on reducing the complexity of the perturbation structure rather than perturbation intensity. To reduce the perturbation complexity, we implement a down-sample operation at the beginning of the dimension reducer $DR$. The perturbation after the down-sample is not the same size as the original samples, which requires the up-sample of the perturbation.

The perturbation after down-sample and up-sample operation is coarse-grained and inaccurate, which would lead to drawbacks in attack ability and increase the training difficulty. To release these drawbacks, we add a skip connection within the dimension reducer $DR$. The effective perturbation outputted by the generator $G$ can skip the down-sample and the up-sample operation through the skip connection. The preservation of this effective perturbation can improve the attack ability. In addition, the skip connection can improve the performance of DNNs, by reducing the training difficulty brought by the deepening of networks \cite{he2016deep}. With the combination of down-sample, up-sample, and skip connection within the dimension reducer $DR$, we effectively reduce the complexity of the perturbation to boost the recoverability of SRAE further. We evaluate the performance of different combinations of down-sample, up-sample, and skip connection in Section \ref{dr}. 

\textbf{\textit{The less complex perturbation is easier to be recovered.}} Given a image $x$ with $n$ pixels, supposing $r=[r_1,r_2,r_3,...,r_n] \sim R^n$ denote the perturbation adding to the each pixel of the image and $r^*=[r^*_1,r^*_2,r^*_3,...,r^*_n] \sim R^n$ denote the recovered perturbation. The adversarial example can be obtained by $x+r$, and the recovered example can be obtained by $x+r-r^*$. Here we use $\Delta$ to represent the difference between $r$ and $r^*$ under $L_2$ norm, which can be calculated as 
\begin{equation}
\begin{aligned}
\Delta &= ||r^*-r||_2\\
&=\sqrt{(r^*_1-r_1)^2+(r^*_2-r_2)^2+...+(r^*_n-r_n)^2}
\end{aligned}
\end{equation}

Let the $r'=[r'_1,r'_2,r'_3,...,r'_n] \sim R^n$ denote the perturbation $r$ pass the dimension reducer $DR$. Correspondingly, the difference $\Delta'$ between $r^*$ and $r'$ can be calculated as 
\begin{equation}
\begin{aligned}
\Delta'&= ||r^*-r'||_2 \\
&=\sqrt{(r^*_1-r'_1)^2+(r^*_2-r'_2)^2+...+(r^*_n-r'_n)^2} 
\end{aligned}
\end{equation} 

For clarity, we take the average pooling with size $m$ (e.g., m=9 for the kernel size is 3$\times$3) as the down-sample and up-sample operation within the dimension reducer $DR$ for illustration, which means
\begin{equation}
r'_1=r'_2=...=r'_m = \overline{r}
\end{equation}
where
\begin{equation}
\overline{r}=\frac{1}{m}\sum_{i=1}^{m}r_i \quad(m \le n)
\end{equation}
The difference between $(\Delta'_m)^2$ and $(\Delta_m)^2$ can be calculated as
\begin{equation}
\begin{aligned}
	(\Delta'_m)^2 - (\Delta_m)^2
	&= \sum_{i=1}^{m}(r^*_i-r'_i)^2 - \sum_{i=1}^{m}(r^*_i-r_i)^2 \\
	&=m\overline{r}^2-m\overline{r^2}	
\end{aligned}
\label{eq1}
\end{equation}
Meanwhile,
\begin{equation}
\begin{aligned}
	\sum_{i=1}^{m}(r_i-\overline{r})^2 &\ge 0 \\
	m\overline{r^2} &\ge m\overline{r}^2
\end{aligned}
\label{eq2}
\end{equation}
Combine the Eq. \ref{eq1} with Eq. \ref{eq2}, it can be obtained that 
\begin{equation}
\Delta' \le \Delta
\label{delta}
\end{equation}
which means $r^*$ is closer to $r'$ (see appendix \ref{proof} for more detail proof). This proves the less complex perturbation is easier to be recovered. As a result, we can further boost the recoverability of SRAE by the proposed dimension reducer $DR$.

\subsection{Loss Function}
\label{loss function}
The loss function for optimizing the generator $G$ can be described as
\begin{equation}
L_G = L_{G\_adv} + \lambda_1L_{G\_dis} + \lambda_2L_{G\_mse}
\end{equation}
where 
\begin{equation}
L_{G\_adv} = H(x+DR(G(x)), y)
\end{equation}
\begin{equation}
L_{G\_dis} = log(1-D(x+DR(G(x))))
\end{equation}
\begin{equation}
L_{G\_mse} = ||DR(G(x))||^2_2
\end{equation}
$||\cdot||_2$ is the $L_2$ norm. $L_{G\_adv}$ aims to improve the attack ability by calculating the cross-entropy loss $H(\cdot)$. $L_{G\_dis}$ aims to make sure the indistinguishability between the generated examples and the original examples. $L_{G\_mse}$ aims to constrain the perturbation intensity. $\lambda_1$ and $\lambda_2$ are the weights of the corresponding losses.

We exploit the $L_{R\_mse}$ to optimize the recover $R$, which can be described as
\begin{equation}
L_{R\_mse} = ||R(x+DR(G(x))) - DR(G(x))||^2_2
\end{equation}
To comprehensively measure the recoverability, we also calculated the adversarial loss of the recovered examples, which can be described as
\begin{equation}
L_{R\_adv} = H(x+DR(G(x))-R(x+DR(G(x))), y)
\end{equation}
We also exploit $L_{R\_adv}$ to to optimize the recover $R$ as competitive solutions (see Section \ref{other loss function for recover}).

For the discriminator $D$, the loss function can be described as: 
\begin{equation}
L_{D} = log(D(x)) + log(1-D(x+DR(G(x))))
\end{equation}
where $log(D(x))$ and $log(1-D(x+DR(G(x))))$ aim to calculate the loss of the original examples and adversarial examples be recognized, respectively.

\section{Experiment and Analysis}
\label{Experiment and Analysis}

For a fair comparison, all experiments are conducted on an NVIDIA RTX 2080Ti, and all methods are implemented by PyTorch. For the attack methods PGD\cite{madry2017towards}, C\&W\cite{cw}, and DDN\cite{rony2019decoupling}, we use the implementation from $advertorch$\cite{ding2019advertorch}. For the defense methods, we use the official implementation of ARN$\footnote{https://github.com/dwDavidxd/ARN}$\cite{zhou2021towards}, APE$\footnote{https://github.com/owruby/APE-GAN}$\cite{jin2019ape}, Image Super-resolution$\footnote{https://github.com/aamir-mustafa/super-resolution-adversarial-defense}$\cite{mustafa2019image}, JPEG Compression$\footnote{https://github.com/poloclub/jpeg-defense}$\cite{das2018shield}, Pixel Deflection$\footnote{https://github.com/iamaaditya/pixel-deflection}$ \cite{prakash2018deflecting}, Random Resizing and Padding$\footnote{https://github.com/cihangxie/NIPS2017\_adv\_challenge\_defense
}$\cite{xie2017mitigating}, Image quilting + total variance minimization$\footnote{https://github.com/facebookresearch/adversarial\_image\_defenses}$\cite{guo2017countering}. All the comparative methods mentioned above are conducted with their default setting. The learning rate for the generator $G$, the recover $R$, and the discriminator $D$ are all set to $10^{-3}$ and decrease $10^{-1}$ times for every 50 epochs (total of 150 epochs). Meanwhile, $\lambda_1 = 10$, $\lambda_2 = 1$ for the $L_G$.

For better measuring the generality, we conduct comparison experiments on various network architectures with different datasets. Specifically, we train LetNet-5 on the MNIST\cite{lecun1998gradient}(the image size is 28$\times$28), and ResNet-50, DenseNet-121, and MoblieNetV3 on Caltech-256 $\footnote{http://www.vision.caltech.edu/Image\_Datasets/Caltech256/}$. Caltech-256\cite{griffin2007caltech} is selected from the Google image dataset. This dataset are divided into 256 categories, with more than 80 images in each category. 

As described in Section \ref{Introduction}, our SRAE aims to serve as a protection mechanism against malicious intelligent detection or classification algorithm. Therefore, our goal is to ensure the target model misclassifies SRAE in various cases (e.g., disturbance, adversarial defense). More importantly, the SRAE should only be able to recover near losslessly by our recover $R$.
\subsection{Ablation Study}

\subsubsection{network structure}
\label{network structure}

\begin{figure}[ht]
	\centering
	\includegraphics[scale=0.25]{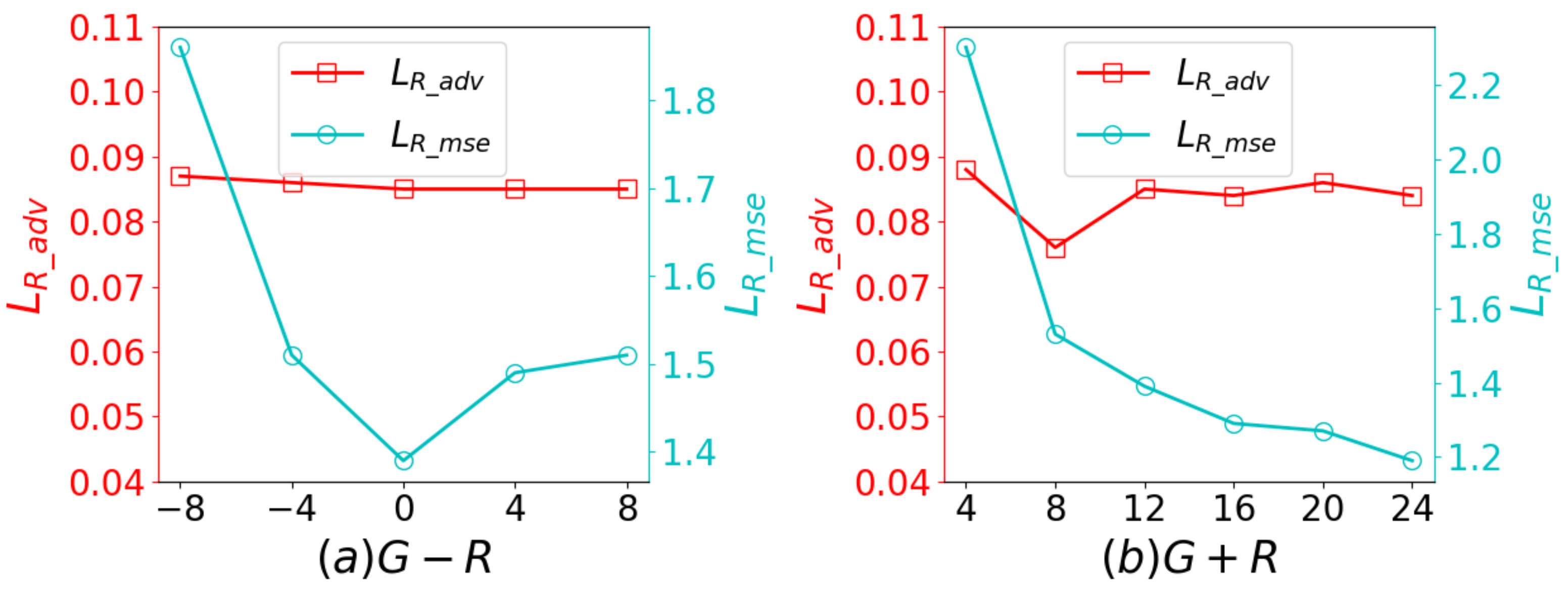}
	\caption{The recoverability of our SRAE with different network depths. (a) shows the recoverability with different bottleneck depths in the generator $G$ and the recover $R$ under a total depth equal to 12. (b) shows the recoverability with the increase of total depth.}
	\label{struct_depth}
\end{figure}

Here, we study the recoverability brought by different depths of networks. To prove the generator $G$ and the recover $R$ with similar depth can improve the recoverability, we train the generator $G$ and the recover $R$ with various depths. As shown in Figure \ref{struct_depth}(a), we fix the total depth of bottleneck in the generator $G$ and the recover $R$ equal to 12. $G-R$ represents the depth difference of the generator $G$ and the recover $R$. From the changes of $L_{R\_mse}$, we can observe that the similar depth of the generator $G$ and the recover $R$ can recover the adversarial examples with fewer errors.

Based on this, we set the generator $G$ and the recover $R$ with the same depth to study the effect of the total depth on the recoverability. As shown in Figure \ref{struct_depth}(b), $G+R$ represents the sum of depth. Although the $L_{R\_adv}$ does not reduce apparently, the $L_{R\_mse}$ keeps reducing with the increase of total depth. It reveals the deeper network can recover the adversarial example with fewer errors. However, to make a trade-off between the parameter quantity and the recoverability, we set the total depth equal to 8 (the depth of both the generator $G$ and the recover $R$ is equal to 4) to conduct the following experiments.

\subsubsection{dimension reducer DR}
\label{dr}
For the dimension reducer $DR$, we first perform an ablation experiment to prove its effectiveness. Then, we focus on the improvement brought by combinations of various down-sample, up-sample, and skip connection operations within the dimension reducer $DR$.

\begin{table}[ht]
	\caption{The comparison between the RGAN without the dimension reducer $DR$ and the RGAN with dimension reducer $DR$. Within dimension reducer $DR$, we further explore the improvement brought by combinations of various down-sample, up-sample, and skip connection operations.}
	\setlength{\tabcolsep}{1.5mm}
	\centering
	\begin{threeparttable} 
		\begin{tabular}{ccccccc}
			\toprule
			\multirow{2}{*}{Skip}              & \multirow{2}{*}{Down} & \multirow{2}{*}{Up} & \multicolumn{2}{c}{Generator}       & \multicolumn{2}{c}{Recover}         \\ \cmidrule(lr){4-5} \cmidrule(lr){6-7}
			&                              &                            & $L_{G\_adv}$              & $L_{G\_mse}$             & $L_{R\_adv}$             & $L_{R\_mse}$             \\ \midrule
			NA\tnote{1}    & NA                        & NA                         & \textbf{0.074}\tnote{2}   & 9.64             & 0.076            & 1.53             \\ \midrule
			\multirow{9}{*}{$\times$\tnote{3}} & \multirow{3}{*}{avg}         & avg                        & 0.188            & 16.24            & 0.029            & 10.51            \\
			&                              & max                        & -\tnote{4}                 & -                & -                & -                \\
			&                              & conv                       & \textbackslash{}\tnote{5}   & \textbackslash{} & \textbackslash{} & \textbackslash{} \\ \cmidrule{2-7} 
			& \multirow{3}{*}{max}         & avg                        & 0.439            & 15.09            & 0.023            & 8.68             \\
			&                              & max                        & 0.309            & 13.92            & 0.088            & 3.2              \\
			&                              & conv                       & \textbackslash{} & \textbackslash{} & \textbackslash{} & \textbackslash{} \\ \cmidrule{2-7} 
			& \multirow{3}{*}{conv}        & avg                        & 0.262            & 11.26            & 0.019            & 6.25             \\
			&                              & max                        & -                & -                & -                & -                \\
			&                              & conv                       & \textbackslash{} & \textbackslash{} & \textbackslash{} & \textbackslash{} \\ \midrule
			\multirow{9}{*}{$\checkmark$}          & \multirow{3}{*}{avg}         & avg                        & 0.139            & \textbf{8.86}    & 0.084            & 1.45             \\
			&                              & max                        & -                & -                & -                & -                \\
			&                              & conv                       & 0.194            & 9.18             & 0.085            & 1.32             \\ \cmidrule{2-7} 
			& \multirow{3}{*}{max}         & avg                        & 0.157            & 8.93             & 0.085            & 1.56             \\
			&                              & max                        & 0.173            & 9.11             & 0.084            & 1.86             \\
			&                              & conv                       & 0.174            & 9.32             & 0.085            & 1.5              \\ \cmidrule{2-7} 
			& \multirow{3}{*}{conv}        & avg                        & 0.201            & \textbf{8.86}    & \textbf{0.008}   & \textbf{1.17}    \\
			&                              & max                        & -                & -                & -                & -                \\
			&                              & conv                       & 0.187            & 8.94             & 0.008            & 1.22             \\ \bottomrule
		\end{tabular}

		\begin{tablenotes}
			\footnotesize
			\item[1] NA represents the RGAN without taking the dimension reducer $DR$.
			\item[2] The best performance was emphasized with \textbf{bold}.
			\item[3] ``$\checkmark$" exploits skip connection within the dimension reducer while ``$\times$" represents without skip connection.
			\item[4] ``-" represents this combination can not be conducted.
			\item[5] ``\textbackslash{}" represents this combination can not converge loss.
		\end{tablenotes} 
	\end{threeparttable} 
	\label{table skip up down}
\end{table}

Table \ref{table skip up down} shows the performance of RGAN without the dimension reducer $DR$, which is represented by NA. Furthermore, Table \ref{table skip up down} also shows the loss of the adversarial examples and the recovered examples generated by RGAN with combinations of various down-sample, up-sample, and skip connection operations. Note that the maximum pooling for up-sample requires the index of the down-sample operation. Thus, only the combination of maximum pooling for both down-sample and up-sample is conducted. From Table \ref{table skip up down}, $L_{G\_adv}$ and $L_{R\_adv}$ reflect the adversarial property of adversarial examples and recovered examples. $L_{G\_mse}$ reflects the error between the adversarial examples and the original examples, while $L_{R\_mse}$ reflects the error between the recovered examples and the original examples.

As shown in Table \ref{table skip up down}, with the comparison of the $L_{R\_adv}$ and $L_{R\_mse}$, the proposed RGAN employing the dimension reducer $DR$ can recover the adversarial examples better than without employing dimension reducer $DR$. Consistent with the discussion in the proposed method, the less complex perturbation is easier to be recovered. From the observation, several combinations in Table \ref{table skip up down}(e.g., convolution for both up-sample and down-sample without skip connection) can not converge the loss. However, by exploiting the skip connection, each combination of up-sample and down-sample can be trained. This proves the skip connection can greatly reduce the training difficulty and make the loss converged better. Furthermore, with the combination of skip connection and convolution for down-sample, the $L_{R\_adv}$ and $L_{R\_mse}$ are less than other combinations. This represents the adversarial property, and the error of the recovered examples both achieve minimization. The advantages prove the convolution is more flexible, which keeps a more effective adversarial perturbation during the down-sample and further boosts the recoverability. 

The column $L_{G\_adv}$ in Table \ref{table skip up down} shows the dimension reducer $DR$ brings some drawbacks in attack ability while improving the recoverability. Therefore, to evaluate the attack ability, we further explore the effect of different reducing levels by setting different kernel sizes with the corresponding stride(e.g., stride = 2 for kernel size = 2$\times$2).

\begin{figure}[h]
	\centering
	\includegraphics[scale=0.65]{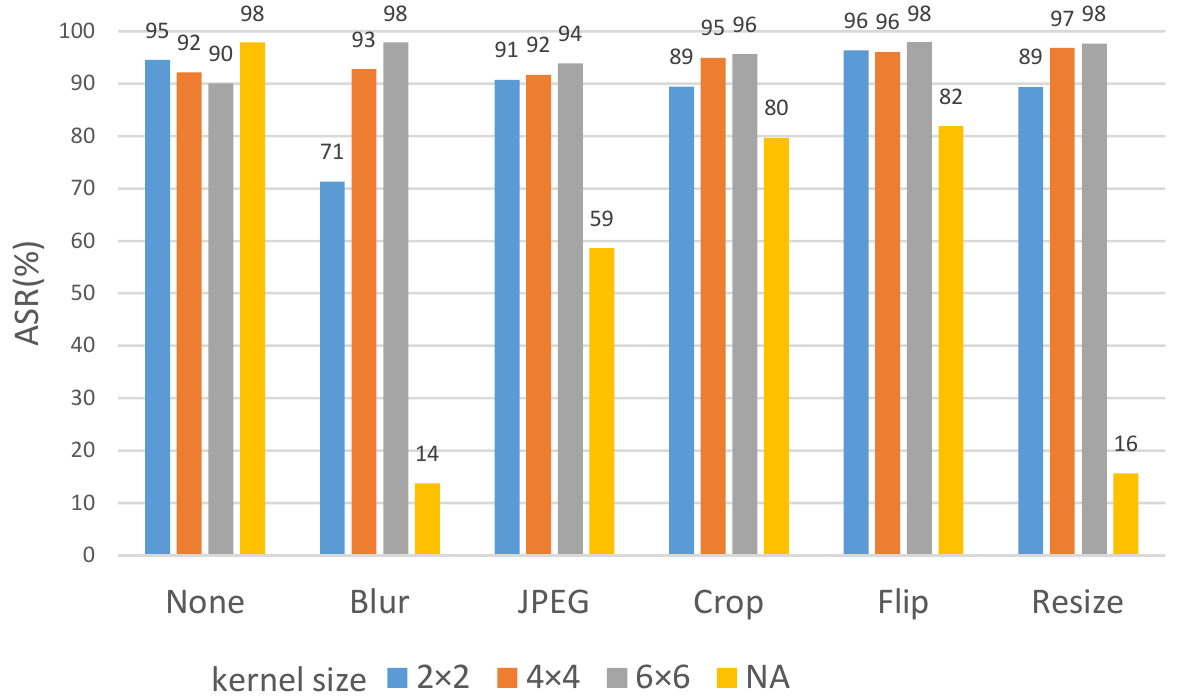}
	\caption{The attack success rate (ASR) of SRAE with various settings. NA represents the RGAN without exploiting the dimension reducer $DR$. In addition to the undisturbed case, we also explore the ASR under the disturbance of widespread image manipulation on social network platforms. From left to right, None represents no transformation, Gaussian blur($\sigma=2$), JPEG compression(quality factor = 70), Center crop(proportion = 90\%), Flip(horizontal and vertical) and Resize(half of the height and width).}
	\label{kernel size}
\end{figure}

We choose the combination of convolution for down-sample and average-pooling for up-sample without skip connection to evaluate the performance of attack ability. As mentioned in Section \ref{Introduction}, our SRAE aims to serve as a protection mechanism on social network platforms. Thus, we also evaluate the robustness of SRAE against the disturbance of widespread image manipulation (e.g., JPEG compression, Gaussian noise) on social network platforms. As shown in Figure \ref{kernel size}, without disturbance, the attack success rate(ASR) of RGAN without employing the dimension reducer $DR$ is 98\%, while exploiting dimension reducer $DR$ can only achieve 90\%-95\%. And a larger kernel size brings lower ASR. It proves that a larger kernel size will lead to a coarser-grained perturbation, resulting in a disadvantage in attack ability. However, as shown in Figure \ref{kernel size}, the coarse-grained perturbation brought by the dimension reducer $DR$ also improves the robustness of SRAE within the disturbing case. Moreover, with the increase in kernel size, the SRAE becomes more and more robust. The robustness against these transformations further ensures the effectiveness and security of the proposed SRAE severed as a protection mechanism in social networks.

\subsubsection{loss function}
\label{other loss function for recover}
 
Here, we study the effect of different loss functions for the recover $R$. As mentioned above, the recoverability can be measured by $L_{R\_adv}$ and $L_{R\_mse}$. We combine the $L_{R\_adv}$ and $L_{R\_mse}$ with different weights $\alpha$ and $\beta$ to form target loss function.
\begin{table}[ht]
	\centering
	\caption{The performance of our RGAN with different loss functions for the recover $R$.}
	\begin{threeparttable} 
		\setlength{\tabcolsep}{1mm}
		\begin{tabular}{ccccc}
			\toprule
			\multirow{2}{*}{$\alpha*L_{R\_mse}+\beta*L_{R\_adv}$} & \multicolumn{2}{c}{Adversarial} & \multicolumn{2}{c}{Recovered}   \\ \cmidrule(lr){2-3} \cmidrule(lr){4-5} 
			& PSNR(dB)     & ACC(\%)     & PSNR(dB)       & ACC(\%)   \\ \midrule
			$\alpha=1,\beta=0$               & 32.21        & 2.10             & \textbf{48.03} \tnote{1} & \textbf{82.76} \\
			$\alpha=0,\beta=1$             & 31.96        & 1.91             & 24.91          & 82.22          \\
			$\alpha=1,\beta=1$         & 32.12        & 2.18             & 29.65          & 82.26          \\
			$\alpha=10,\beta=1$      & 31.95        & 1.79             & 36.15          & 81.91          \\
			$\alpha=100,\beta=1$     & 32.11        & 2.52             & 40.35          & 82.02          \\
			\bottomrule
		\end{tabular}
		\begin{tablenotes}
			\footnotesize
			\item[1] The best performance of recoverability was emphasized with \textbf{bold}.  
		\end{tablenotes} 
	\end{threeparttable} 
		\label{loss}
\end{table}

PSNR within the column adversarial reflects the difference between adversarial and original example, while PSNR within the column recovered reflects the difference between recovered and original example. ACC reflects the classification accuracy for adversarial and recovered examples. As shown in Table \ref{loss}, only taking $L_{R\_adv}$ ($\alpha=0,\beta=1$) as the loss function can not achieve satisfactory recoverability. The difference between the recovered and original examples is even larger than the difference between the adversarial and original examples. In addition, we try to alleviate between $L_{R\_mse}$ and $L_{R\_adv}$ by setting different $\alpha$ and $\beta$. However, the improvement of PSNR can not be consistent with improving ACC. These phenomena partly reveal the defects of the decision boundary of the target network. Recovering the examples according to this decision boundary may enlarge the difference between the recovered and original examples. Thus, we only exploit $L_{R\_mse}$ for the optimization, which is superior in both difference and recoverability.

\subsection{Recoverability Comparison}
\begin{table*}[h]
	\caption{Recoverability comparison between state-of-the-art defense against DDN attack and our RGAN on MNIST and Caltech-256. Our method achieves better recoverability across different datasets and different network architectures.}
	\setlength{\tabcolsep}{1.2mm}
	\begin{threeparttable} 
	\begin{tabular}{ccccccccccccc}
		\toprule
		& \multicolumn{3}{c}{MNIST}                        & \multicolumn{9}{c}{Caltech-256}                                                                                                                           \\ \cmidrule(lr){2-4} \cmidrule(lr){5-13} 
		& \multicolumn{3}{c}{LeNet-5}                      & \multicolumn{3}{c}{DenseNet-121}                  & \multicolumn{3}{c}{ResNet-50}                     & \multicolumn{3}{c}{MoblieNetV3}                   \\ \cmidrule(lr){2-4} \cmidrule(lr){5-7} \cmidrule(lr){8-10} \cmidrule(lr){11-13}
		& $L_2$            & PSNR(dB)            & CER(\%) \tnote{1}     & $L_2$            & PSNR(dB)             & CER(\%)         & $L_2$            & PSNR(dB)             & CER(\%)         & $L_2$            & PSNR(dB)             & CER(\%)         \\ \midrule
		Xie $et$   $al.$\cite{xie2017mitigating}         & 7.66          & 11.60            & 43.74         & 89.19         & 13.54            & 30.66          & 88.71         & 13.59            & 28.05          & 88.82         & 13.55            & 35.75          \\
		Guo $et$   $al.$\cite{guo2017countering}         & 3.02          & 19.45            & 26.58         & 11.26         & 30.57            & 27.70          & 11.26         & 30.56            & 27.82          & 11.26         & 30.59            & 35.37          \\
		Prakash $et$   $al.$\cite{prakash2018deflecting} & 4.43          & 16.13            & 50.06         & 8.85          & 32.91            & 24.04          & 8.84          & 32.91            & 24.35          & 8.84          & 32.91            & 29.80          \\
		Das $et$   $al.$\cite{das2018shield}             & 1.67          & 24.78            & 37.07         & 7.77          & 34.49            & 20.89          & 7.78          & 34.48            & 21.47          & 7.77          & 34.49            & 26.22          \\
		Mustafa $et$   $al.$\cite{mustafa2019image}      & 2.00          & 23.26            & 19.78         & 11.04         & 31.04            & 26.92          & 11.04         & 31.04            & 25.13          & 11.04         & 31.04            & 30.11          \\
		Jin  $et$ $al.$\cite{jin2019ape}                 & 1.65          & 24.73            & 1.73          & 24.53         & 24.03            & 25.33          & 20.56         & 25.85            & 26.88          & 20.65         & 25.57            & 35.79          \\
		Zhou  $et$ $al.$\cite{zhou2021towards}           & 1.11          & 28.24            & 1.45          & 39.38         & 20.34            & 93.85          & 39.45         & 20.36            & 93.34          & 38.59         & 20.57            & 92.33          \\
		RGAN(our)                                                         & \textbf{0.50}\tnote{2} & \textbf{35.24}   & \textbf{1.16} & \textbf{1.29} & \textbf{48.09}   & \textbf{17.19} & \textbf{2.44} & \textbf{43.73}   & \textbf{16.92} & \textbf{1.84} & \textbf{45.97}   & \textbf{21.36} \\ \midrule
		Clean                                                                & 0.00          & $+\infty$ & 1.11          & 0.00          & $+\infty$ & 17.00          & 0.00          & $+\infty$ & 16.34          & 0.00          & $+\infty$ & 20.85          \\ \bottomrule
	\end{tabular}

	\begin{tablenotes}
		\footnotesize
		\item[1] CER represent the classification error rate (\%) of the target network. 
		\item[2] The best performance was emphasized with \textbf{bold}. 
	\end{tablenotes} 
	\end{threeparttable} 
	\label{defense attack ddn}
\end{table*}

Table \ref{defense attack ddn} shows the recoverability results on different datasets with various network architectures. Due to the lack of recoverability study of adversarial examples, we combine the classical DDN \cite{rony2019decoupling} with state-of-the-art adversarial recovery or defense methods to serve as competitive solutions with our SRAE. We also compare C\&W \cite{cw} and PGD \cite{madry2017towards} in the appendix \ref{other recoverability}. The recoverability is measured from two aspects: the difference between the recovered and original examples and the adversarial property of the recovered examples. Specifically, the difference is reflected by the two criteria: the $L_2$ norm of the error (smaller is better) and peak signal to noise ratio (PSNR, larger is better). And the adversarial property is reflected by the classification error rate (CER, lower is better) of the target network towards these recovered examples. From Table \ref{defense attack ddn}, minor difference and lower CER represent better recoverability of the solution. It can be observed that our RGAN consistently achieves the best recoverability on both the small size MNIST (with image size 28$\times$28) and the large size Caltech-256 (with image size 224$\times$224) with various network architectures. This advantage ensures the effectiveness and generality of our RGAN served as a protection mechanism.

On MNIST, the combination of DDN for attack with several defense methods(e.g., \cite{zhou2021towards}, \cite{jin2019ape}, \cite{mustafa2019image}, \cite{das2018shield}) shows competitive performance, recovering around 98\% of the adversarial examples. In comparison, our RGAN recovers near 99\% of the examples with much more minor errors($L_2$ norm reduce more than halved). Besides, our RGAN can maintain the recoverability (still can recover near 99\% of the examples with a small error) when transferring to larger datasets and other network structures. However, the above competitive combinations on MNIST don't show a satisfactory transfer of recoverability. 

On Caltech-256, the combination of DDN and Das $et$ $al.$\cite{das2018shield} outperform other combinations in recoverability. In comparison, our RGAN still recovers 3\% of the examples more than this combination. Note that the defense method proposed by Das $et$ $al.$\cite{das2018shield} are based on JPEG Compression. This means that Das $et$ $al.$\cite{das2018shield} reduces the CER based on the destruction rather than the recovery of adversarial perturbation, which can be reflected by the larger errors between the recovered and original examples (larger $L_2$ norm and smaller PSNR). Based on this consideration, these combinations are unsuitable for serving as a protection mechanism.

\subsection{Adaptive Attack Evaluation}

\begin{figure}[h]
	\centering
	\includegraphics[scale=0.5]{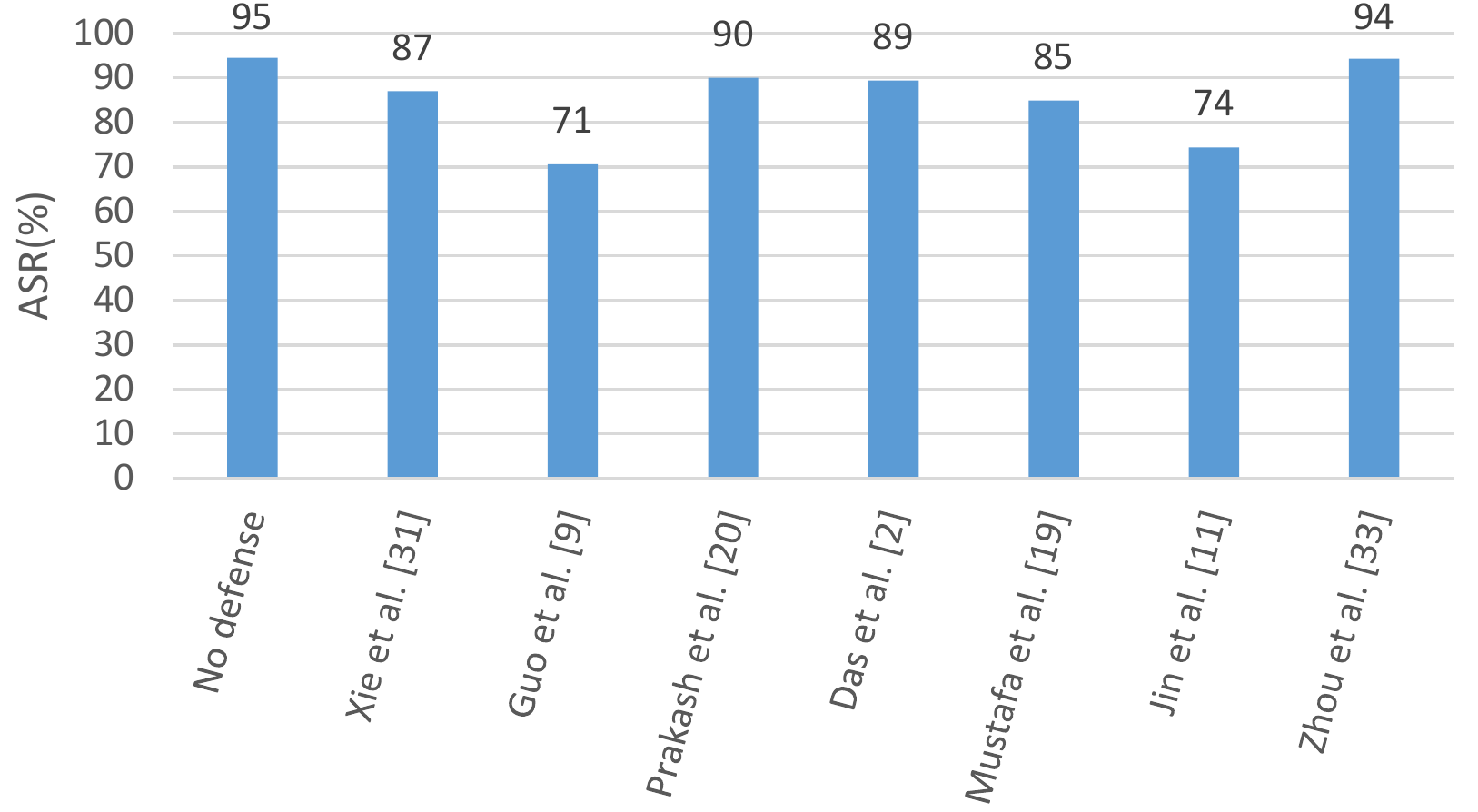}
	\caption{The adaptive attack ability of our SRAE. ASR represents the attack success rate of the target DNNs after being attacked by our SRAE. From left to right, we evaluate the attack ability of our SRAE against various state-of-the-art adversarial defense methods.}
	\label{adaptive attack eva}
\end{figure}

As described in Section \ref{Introduction}, SRAE aims to serve as a new effective protection mechanism against malicious intelligent detection algorithms in social network platforms. To more comprehensively measure the effectiveness of this protection mechanism, we evaluate the attack ability under different disturbance conditions in Figure \ref{kernel size}. Apart from disturbance conditions, if the attackers are aware of our adversarial protection mechanism, can they destroy our SRAE by exploiting the existing adversarial defense methods? Thus, we further evaluate the attack ability of SRAE under the state-of-the-art adversarial defense methods. 

As shown in Figure \ref{adaptive attack eva}, the attack success rate of our SRAE achieves 95\% without adversarial defense. In addition, under various adversarial defense strategies, the SRAE still achieves a high attack success rate (from 71\% to 94\%). It can be observed from Figure \ref{kernel size} and Figure \ref{adaptive attack eva}, our SRAE can maintain high attack ability in both interference and adversarial defense conditions, which proves the SRAE can serve as an effective and robust protection mechanism for users' privacy in social network platforms.

\section{Conclusion}
Despite the destruction and threat brought by the adversarial examples, we transformed these negative effects into a positive protection mechanism in this paper. We proposed a $recoverable$ $generative$ $adversarial$ $network$ (RGAN) to generate $self$-$recoverable$ $advers$-\\$arial$ $examples$ (SRAE). Specifically, by joint dynamic training of the generator $G$ and the recover $R$, the proposed model improved the recoverability while maintaining the attack ability. Besides, by studying the effects of perturbation intensity and complexity on the recoverability, we designed a dimension reducer $DR$ to optimize the perturbation distribution and further boost the recoverability. Experimental results demonstrated that our model presented superior recoverability than the combinations of state-of-the-art attack and defense methods on different datasets and network architectures. The advantage in recoverability, attack ability, and robustness ensures the effectiveness and generality of our model served as a new effective protection mechanism for privacy security in social networks. 

Since the proposed RGAN is lightweight, in future work, we will try to explore the upper limit for recoverability by training a deeper and wider network architecture with more data.

\bibliographystyle{ACM-Reference-Format}
\bibliography{ref}

%%% -*-BibTeX-*-
%%% Do NOT edit. File created by BibTeX with style
%%% ACM-Reference-Format-Journals [18-Jan-2012].

\begin{thebibliography}{35}

%%% ====================================================================
%%% NOTE TO THE USER: you can override these defaults by providing
%%% customized versions of any of these macros before the \bibliography
%%% command.  Each of them MUST provide its own final punctuation,
%%% except for \shownote{}, \showDOI{}, and \showURL{}.  The latter two
%%% do not use final punctuation, in order to avoid confusing it with
%%% the Web address.
%%%
%%% To suppress output of a particular field, define its macro to expand
%%% to an empty string, or better, \unskip, like this:
%%%
%%% \newcommand{\showDOI}[1]{\unskip}   % LaTeX syntax
%%%
%%% \def \showDOI #1{\unskip}           % plain TeX syntax
%%%
%%% ====================================================================

\ifx \showCODEN    \undefined \def \showCODEN     #1{\unskip}     \fi
\ifx \showDOI      \undefined \def \showDOI       #1{#1}\fi
\ifx \showISBNx    \undefined \def \showISBNx     #1{\unskip}     \fi
\ifx \showISBNxiii \undefined \def \showISBNxiii  #1{\unskip}     \fi
\ifx \showISSN     \undefined \def \showISSN      #1{\unskip}     \fi
\ifx \showLCCN     \undefined \def \showLCCN      #1{\unskip}     \fi
\ifx \shownote     \undefined \def \shownote      #1{#1}          \fi
\ifx \showarticletitle \undefined \def \showarticletitle #1{#1}   \fi
\ifx \showURL      \undefined \def \showURL       {\relax}        \fi
% The following commands are used for tagged output and should be
% invisible to TeX
\providecommand\bibfield[2]{#2}
\providecommand\bibinfo[2]{#2}
\providecommand\natexlab[1]{#1}
\providecommand\showeprint[2][]{arXiv:#2}

\bibitem[\protect\citeauthoryear{Carlini and Wagner}{Carlini and
  Wagner}{2017}]%
        {cw}
\bibfield{author}{\bibinfo{person}{Nicholas Carlini} {and}
  \bibinfo{person}{David Wagner}.} \bibinfo{year}{2017}\natexlab{}.
\newblock \bibinfo{title}{Towards Evaluating the Robustness of Neural
  Networks}.
\newblock
\newblock
\showeprint[arxiv]{1608.04644}~[cs.CR]


\bibitem[\protect\citeauthoryear{Cheng, Dong, Pang, Su, and Zhu}{Cheng
  et~al\mbox{.}}{2019}]%
        {cheng2019improving}
\bibfield{author}{\bibinfo{person}{Shuyu Cheng}, \bibinfo{person}{Yinpeng
  Dong}, \bibinfo{person}{Tianyu Pang}, \bibinfo{person}{Hang Su}, {and}
  \bibinfo{person}{Jun Zhu}.} \bibinfo{year}{2019}\natexlab{}.
\newblock \showarticletitle{Improving black-box adversarial attacks with a
  transfer-based prior}.
\newblock \bibinfo{journal}{\emph{Advances in neural information processing
  systems}}  \bibinfo{volume}{32} (\bibinfo{year}{2019}).
\newblock


\bibitem[\protect\citeauthoryear{Das, Shanbhogue, Chen, Hohman, Li, Chen,
  Kounavis, and Chau}{Das et~al\mbox{.}}{2018}]%
        {das2018shield}
\bibfield{author}{\bibinfo{person}{Nilaksh Das}, \bibinfo{person}{Madhuri
  Shanbhogue}, \bibinfo{person}{Shang-Tse Chen}, \bibinfo{person}{Fred Hohman},
  \bibinfo{person}{Siwei Li}, \bibinfo{person}{Li Chen},
  \bibinfo{person}{Michael~E Kounavis}, {and} \bibinfo{person}{Duen~Horng
  Chau}.} \bibinfo{year}{2018}\natexlab{}.
\newblock \showarticletitle{Shield: Fast, practical defense and vaccination for
  deep learning using jpeg compression}. In
  \bibinfo{booktitle}{\emph{Proceedings of the 24th ACM SIGKDD International
  Conference on Knowledge Discovery \& Data Mining}}.
  \bibinfo{pages}{196--204}.
\newblock


\bibitem[\protect\citeauthoryear{Ding, Wang, and Jin}{Ding
  et~al\mbox{.}}{2019}]%
        {ding2019advertorch}
\bibfield{author}{\bibinfo{person}{Gavin~Weiguang Ding}, \bibinfo{person}{Luyu
  Wang}, {and} \bibinfo{person}{Xiaomeng Jin}.}
  \bibinfo{year}{2019}\natexlab{}.
\newblock \showarticletitle{{AdverTorch} v0.1: An Adversarial Robustness
  Toolbox based on PyTorch}.
\newblock \bibinfo{journal}{\emph{arXiv preprint arXiv:1902.07623}}
  (\bibinfo{year}{2019}).
\newblock


\bibitem[\protect\citeauthoryear{Dong, Liao, Pang, Su, Zhu, Hu, and Li}{Dong
  et~al\mbox{.}}{2018}]%
        {dong2018boosting}
\bibfield{author}{\bibinfo{person}{Yinpeng Dong}, \bibinfo{person}{Fangzhou
  Liao}, \bibinfo{person}{Tianyu Pang}, \bibinfo{person}{Hang Su},
  \bibinfo{person}{Jun Zhu}, \bibinfo{person}{Xiaolin Hu}, {and}
  \bibinfo{person}{Jianguo Li}.} \bibinfo{year}{2018}\natexlab{}.
\newblock \showarticletitle{Boosting adversarial attacks with momentum}. In
  \bibinfo{booktitle}{\emph{Proceedings of the IEEE conference on computer
  vision and pattern recognition}}. \bibinfo{pages}{9185--9193}.
\newblock


\bibitem[\protect\citeauthoryear{Gong, Wang, and Ku}{Gong
  et~al\mbox{.}}{2017}]%
        {gong2017adversarial}
\bibfield{author}{\bibinfo{person}{Zhitao Gong}, \bibinfo{person}{Wenlu Wang},
  {and} \bibinfo{person}{Wei-Shinn Ku}.} \bibinfo{year}{2017}\natexlab{}.
\newblock \showarticletitle{Adversarial and clean data are not twins}.
\newblock \bibinfo{journal}{\emph{arXiv preprint arXiv:1704.04960}}
  (\bibinfo{year}{2017}).
\newblock


\bibitem[\protect\citeauthoryear{Goodfellow, Shlens, and Szegedy}{Goodfellow
  et~al\mbox{.}}{2015}]%
        {fgsm}
\bibfield{author}{\bibinfo{person}{Ian~J. Goodfellow},
  \bibinfo{person}{Jonathon Shlens}, {and} \bibinfo{person}{Christian
  Szegedy}.} \bibinfo{year}{2015}\natexlab{}.
\newblock \bibinfo{title}{Explaining and Harnessing Adversarial Examples}.
\newblock
\newblock
\showeprint[arxiv]{1412.6572}~[stat.ML]


\bibitem[\protect\citeauthoryear{Griffin, Holub, and Perona}{Griffin
  et~al\mbox{.}}{2007}]%
        {griffin2007caltech}
\bibfield{author}{\bibinfo{person}{Gregory Griffin}, \bibinfo{person}{Alex
  Holub}, {and} \bibinfo{person}{Pietro Perona}.}
  \bibinfo{year}{2007}\natexlab{}.
\newblock \showarticletitle{Caltech-256 object category dataset}.
\newblock  (\bibinfo{year}{2007}).
\newblock


\bibitem[\protect\citeauthoryear{Grosse, Manoharan, Papernot, Backes, and
  McDaniel}{Grosse et~al\mbox{.}}{2017}]%
        {grosse2017statistical}
\bibfield{author}{\bibinfo{person}{Kathrin Grosse}, \bibinfo{person}{Praveen
  Manoharan}, \bibinfo{person}{Nicolas Papernot}, \bibinfo{person}{Michael
  Backes}, {and} \bibinfo{person}{Patrick McDaniel}.}
  \bibinfo{year}{2017}\natexlab{}.
\newblock \showarticletitle{On the (statistical) detection of adversarial
  examples}.
\newblock \bibinfo{journal}{\emph{arXiv preprint arXiv:1702.06280}}
  (\bibinfo{year}{2017}).
\newblock


\bibitem[\protect\citeauthoryear{Guo, Rana, Cisse, and Van Der~Maaten}{Guo
  et~al\mbox{.}}{2017}]%
        {guo2017countering}
\bibfield{author}{\bibinfo{person}{Chuan Guo}, \bibinfo{person}{Mayank Rana},
  \bibinfo{person}{Moustapha Cisse}, {and} \bibinfo{person}{Laurens Van
  Der~Maaten}.} \bibinfo{year}{2017}\natexlab{}.
\newblock \showarticletitle{Countering adversarial images using input
  transformations}.
\newblock \bibinfo{journal}{\emph{arXiv preprint arXiv:1711.00117}}
  (\bibinfo{year}{2017}).
\newblock


\bibitem[\protect\citeauthoryear{He, Zhang, Ren, and Sun}{He
  et~al\mbox{.}}{2016}]%
        {he2016deep}
\bibfield{author}{\bibinfo{person}{Kaiming He}, \bibinfo{person}{Xiangyu
  Zhang}, \bibinfo{person}{Shaoqing Ren}, {and} \bibinfo{person}{Jian Sun}.}
  \bibinfo{year}{2016}\natexlab{}.
\newblock \showarticletitle{Deep residual learning for image recognition}. In
  \bibinfo{booktitle}{\emph{Proceedings of the IEEE conference on computer
  vision and pattern recognition}}. \bibinfo{pages}{770--778}.
\newblock


\bibitem[\protect\citeauthoryear{Jin, Shen, Zhang, Dai, and Zhang}{Jin
  et~al\mbox{.}}{2019}]%
        {jin2019ape}
\bibfield{author}{\bibinfo{person}{Guoqing Jin}, \bibinfo{person}{Shiwei Shen},
  \bibinfo{person}{Dongming Zhang}, \bibinfo{person}{Feng Dai}, {and}
  \bibinfo{person}{Yongdong Zhang}.} \bibinfo{year}{2019}\natexlab{}.
\newblock \showarticletitle{Ape-gan: Adversarial perturbation elimination with
  gan}. In \bibinfo{booktitle}{\emph{ICASSP 2019-2019 IEEE International
  Conference on Acoustics, Speech and Signal Processing (ICASSP)}}. IEEE,
  \bibinfo{pages}{3842--3846}.
\newblock


\bibitem[\protect\citeauthoryear{Kurakin, Goodfellow, and Bengio}{Kurakin
  et~al\mbox{.}}{2017}]%
        {ifgsm}
\bibfield{author}{\bibinfo{person}{Alexey Kurakin}, \bibinfo{person}{Ian
  Goodfellow}, {and} \bibinfo{person}{Samy Bengio}.}
  \bibinfo{year}{2017}\natexlab{}.
\newblock \bibinfo{title}{Adversarial examples in the physical world}.
\newblock
\newblock
\showeprint[arxiv]{1607.02533}~[cs.CV]


\bibitem[\protect\citeauthoryear{LeCun, Bottou, Bengio, and Haffner}{LeCun
  et~al\mbox{.}}{1998}]%
        {lecun1998gradient}
\bibfield{author}{\bibinfo{person}{Yann LeCun}, \bibinfo{person}{L{\'e}on
  Bottou}, \bibinfo{person}{Yoshua Bengio}, {and} \bibinfo{person}{Patrick
  Haffner}.} \bibinfo{year}{1998}\natexlab{}.
\newblock \showarticletitle{Gradient-based learning applied to document
  recognition}.
\newblock \bibinfo{journal}{\emph{Proc. IEEE}} \bibinfo{volume}{86},
  \bibinfo{number}{11} (\bibinfo{year}{1998}), \bibinfo{pages}{2278--2324}.
\newblock


\bibitem[\protect\citeauthoryear{Liang, Li, Su, Li, Shi, and Wang}{Liang
  et~al\mbox{.}}{2018}]%
        {liang2018detecting}
\bibfield{author}{\bibinfo{person}{Bin Liang}, \bibinfo{person}{Hongcheng Li},
  \bibinfo{person}{Miaoqiang Su}, \bibinfo{person}{Xirong Li},
  \bibinfo{person}{Wenchang Shi}, {and} \bibinfo{person}{Xiaofeng Wang}.}
  \bibinfo{year}{2018}\natexlab{}.
\newblock \showarticletitle{Detecting adversarial image examples in deep neural
  networks with adaptive noise reduction}.
\newblock \bibinfo{journal}{\emph{IEEE Transactions on Dependable and Secure
  Computing}} (\bibinfo{year}{2018}).
\newblock


\bibitem[\protect\citeauthoryear{Liao, Liang, Dong, Pang, Hu, and Zhu}{Liao
  et~al\mbox{.}}{2018}]%
        {liao2018defense}
\bibfield{author}{\bibinfo{person}{Fangzhou Liao}, \bibinfo{person}{Ming
  Liang}, \bibinfo{person}{Yinpeng Dong}, \bibinfo{person}{Tianyu Pang},
  \bibinfo{person}{Xiaolin Hu}, {and} \bibinfo{person}{Jun Zhu}.}
  \bibinfo{year}{2018}\natexlab{}.
\newblock \showarticletitle{Defense against adversarial attacks using
  high-level representation guided denoiser}. In
  \bibinfo{booktitle}{\emph{Proceedings of the IEEE Conference on Computer
  Vision and Pattern Recognition}}. \bibinfo{pages}{1778--1787}.
\newblock


\bibitem[\protect\citeauthoryear{Liu, Zhang, Zhang, Hou, Liu, Zha, and Yu}{Liu
  et~al\mbox{.}}{2019}]%
        {liu2019detection}
\bibfield{author}{\bibinfo{person}{Jiayang Liu}, \bibinfo{person}{Weiming
  Zhang}, \bibinfo{person}{Yiwei Zhang}, \bibinfo{person}{Dongdong Hou},
  \bibinfo{person}{Yujia Liu}, \bibinfo{person}{Hongyue Zha}, {and}
  \bibinfo{person}{Nenghai Yu}.} \bibinfo{year}{2019}\natexlab{}.
\newblock \showarticletitle{Detection based defense against adversarial
  examples from the steganalysis point of view}. In
  \bibinfo{booktitle}{\emph{Proceedings of the IEEE/CVF Conference on Computer
  Vision and Pattern Recognition}}. \bibinfo{pages}{4825--4834}.
\newblock


\bibitem[\protect\citeauthoryear{Liu and Hsieh}{Liu and Hsieh}{2019}]%
        {Liu_2019_CVPR}
\bibfield{author}{\bibinfo{person}{Xuanqing Liu} {and} \bibinfo{person}{Cho-Jui
  Hsieh}.} \bibinfo{year}{2019}\natexlab{}.
\newblock \showarticletitle{Rob-GAN: Generator, Discriminator, and Adversarial
  Attacker}. In \bibinfo{booktitle}{\emph{Proceedings of the IEEE/CVF
  Conference on Computer Vision and Pattern Recognition (CVPR)}}.
\newblock


\bibitem[\protect\citeauthoryear{Madry, Makelov, Schmidt, Tsipras, and
  Vladu}{Madry et~al\mbox{.}}{2017}]%
        {madry2017towards}
\bibfield{author}{\bibinfo{person}{Aleksander Madry},
  \bibinfo{person}{Aleksandar Makelov}, \bibinfo{person}{Ludwig Schmidt},
  \bibinfo{person}{Dimitris Tsipras}, {and} \bibinfo{person}{Adrian Vladu}.}
  \bibinfo{year}{2017}\natexlab{}.
\newblock \showarticletitle{Towards deep learning models resistant to
  adversarial attacks}.
\newblock \bibinfo{journal}{\emph{arXiv preprint arXiv:1706.06083}}
  (\bibinfo{year}{2017}).
\newblock


\bibitem[\protect\citeauthoryear{Mustafa, Khan, Hayat, Shen, and Shao}{Mustafa
  et~al\mbox{.}}{2019}]%
        {mustafa2019image}
\bibfield{author}{\bibinfo{person}{Aamir Mustafa}, \bibinfo{person}{Salman~H
  Khan}, \bibinfo{person}{Munawar Hayat}, \bibinfo{person}{Jianbing Shen},
  {and} \bibinfo{person}{Ling Shao}.} \bibinfo{year}{2019}\natexlab{}.
\newblock \showarticletitle{Image super-resolution as a defense against
  adversarial attacks}.
\newblock \bibinfo{journal}{\emph{IEEE Transactions on Image Processing}}
  \bibinfo{volume}{29} (\bibinfo{year}{2019}), \bibinfo{pages}{1711--1724}.
\newblock


\bibitem[\protect\citeauthoryear{Prakash, Moran, Garber, DiLillo, and
  Storer}{Prakash et~al\mbox{.}}{2018}]%
        {prakash2018deflecting}
\bibfield{author}{\bibinfo{person}{Aaditya Prakash}, \bibinfo{person}{Nick
  Moran}, \bibinfo{person}{Solomon Garber}, \bibinfo{person}{Antonella
  DiLillo}, {and} \bibinfo{person}{James Storer}.}
  \bibinfo{year}{2018}\natexlab{}.
\newblock \showarticletitle{Deflecting adversarial attacks with pixel
  deflection}. In \bibinfo{booktitle}{\emph{Proceedings of the IEEE conference
  on computer vision and pattern recognition}}. \bibinfo{pages}{8571--8580}.
\newblock


\bibitem[\protect\citeauthoryear{Rony, Hafemann, Oliveira, Ayed, Sabourin, and
  Granger}{Rony et~al\mbox{.}}{2019}]%
        {rony2019decoupling}
\bibfield{author}{\bibinfo{person}{J{\'e}r{\^o}me Rony},
  \bibinfo{person}{Luiz~G Hafemann}, \bibinfo{person}{Luiz~S Oliveira},
  \bibinfo{person}{Ismail~Ben Ayed}, \bibinfo{person}{Robert Sabourin}, {and}
  \bibinfo{person}{Eric Granger}.} \bibinfo{year}{2019}\natexlab{}.
\newblock \showarticletitle{Decoupling direction and norm for efficient
  gradient-based l2 adversarial attacks and defenses}. In
  \bibinfo{booktitle}{\emph{Proceedings of the IEEE/CVF Conference on Computer
  Vision and Pattern Recognition}}. \bibinfo{pages}{4322--4330}.
\newblock


\bibitem[\protect\citeauthoryear{Szegedy, Zaremba, Sutskever, Bruna, Erhan,
  Goodfellow, and Fergus}{Szegedy et~al\mbox{.}}{2014}]%
        {lbfgs}
\bibfield{author}{\bibinfo{person}{Christian Szegedy},
  \bibinfo{person}{Wojciech Zaremba}, \bibinfo{person}{Ilya Sutskever},
  \bibinfo{person}{Joan Bruna}, \bibinfo{person}{Dumitru Erhan},
  \bibinfo{person}{Ian Goodfellow}, {and} \bibinfo{person}{Rob Fergus}.}
  \bibinfo{year}{2014}\natexlab{}.
\newblock \bibinfo{title}{Intriguing properties of neural networks}.
\newblock
\newblock
\showeprint[arxiv]{1312.6199}~[cs.CV]


\bibitem[\protect\citeauthoryear{Tang, Gong, Wang, Liu, Wang, Chen, Yu, Liu,
  Song, Yuille, et~al\mbox{.}}{Tang et~al\mbox{.}}{2021}]%
        {tang2021robustart}
\bibfield{author}{\bibinfo{person}{Shiyu Tang}, \bibinfo{person}{Ruihao Gong},
  \bibinfo{person}{Yan Wang}, \bibinfo{person}{Aishan Liu},
  \bibinfo{person}{Jiakai Wang}, \bibinfo{person}{Xinyun Chen},
  \bibinfo{person}{Fengwei Yu}, \bibinfo{person}{Xianglong Liu},
  \bibinfo{person}{Dawn Song}, \bibinfo{person}{Alan Yuille}, {et~al\mbox{.}}}
  \bibinfo{year}{2021}\natexlab{}.
\newblock \showarticletitle{Robustart: Benchmarking robustness on architecture
  design and training techniques}.
\newblock \bibinfo{journal}{\emph{arXiv preprint arXiv:2109.05211}}
  (\bibinfo{year}{2021}).
\newblock


\bibitem[\protect\citeauthoryear{Tonge}{Tonge}{2018}]%
        {tonge2018identifying}
\bibfield{author}{\bibinfo{person}{Ashwini Tonge}.}
  \bibinfo{year}{2018}\natexlab{}.
\newblock \showarticletitle{Identifying private content for online image
  sharing}. In \bibinfo{booktitle}{\emph{Proceedings of the AAAI Conference on
  Artificial Intelligence}}, Vol.~\bibinfo{volume}{32}.
\newblock


\bibitem[\protect\citeauthoryear{Tonge and Caragea}{Tonge and Caragea}{2020}]%
        {tonge2020image}
\bibfield{author}{\bibinfo{person}{Ashwini Tonge} {and}
  \bibinfo{person}{Cornelia Caragea}.} \bibinfo{year}{2020}\natexlab{}.
\newblock \showarticletitle{Image privacy prediction using deep neural
  networks}.
\newblock \bibinfo{journal}{\emph{ACM Transactions on the Web (TWEB)}}
  \bibinfo{volume}{14}, \bibinfo{number}{2} (\bibinfo{year}{2020}),
  \bibinfo{pages}{1--32}.
\newblock


\bibitem[\protect\citeauthoryear{Tram{\`e}r, Kurakin, Papernot, Goodfellow,
  Boneh, and McDaniel}{Tram{\`e}r et~al\mbox{.}}{2017}]%
        {tramer2017ensemble}
\bibfield{author}{\bibinfo{person}{Florian Tram{\`e}r}, \bibinfo{person}{Alexey
  Kurakin}, \bibinfo{person}{Nicolas Papernot}, \bibinfo{person}{Ian
  Goodfellow}, \bibinfo{person}{Dan Boneh}, {and} \bibinfo{person}{Patrick
  McDaniel}.} \bibinfo{year}{2017}\natexlab{}.
\newblock \showarticletitle{Ensemble adversarial training: Attacks and
  defenses}.
\newblock \bibinfo{journal}{\emph{arXiv preprint arXiv:1705.07204}}
  (\bibinfo{year}{2017}).
\newblock


\bibitem[\protect\citeauthoryear{Wang, Zhao, Yin, Luo, Zheng, Shi, and
  Jha}{Wang et~al\mbox{.}}{2021}]%
        {wang2021smsnet}
\bibfield{author}{\bibinfo{person}{Jinwei Wang}, \bibinfo{person}{Junjie Zhao},
  \bibinfo{person}{Qilin Yin}, \bibinfo{person}{Xiangyang Luo},
  \bibinfo{person}{Yuhui Zheng}, \bibinfo{person}{Yun~Qing Shi}, {and}
  \bibinfo{person}{Sunil~K Jha}.} \bibinfo{year}{2021}\natexlab{}.
\newblock \showarticletitle{SmsNet: A New Deep Convolutional Neural Network
  Model for Adversarial Example Detection}.
\newblock \bibinfo{journal}{\emph{IEEE Transactions on Multimedia}}
  (\bibinfo{year}{2021}).
\newblock


\bibitem[\protect\citeauthoryear{Wu, Xia, and Wang}{Wu et~al\mbox{.}}{2020}]%
        {wu2020adversarial}
\bibfield{author}{\bibinfo{person}{Dongxian Wu}, \bibinfo{person}{Shu-Tao Xia},
  {and} \bibinfo{person}{Yisen Wang}.} \bibinfo{year}{2020}\natexlab{}.
\newblock \showarticletitle{Adversarial weight perturbation helps robust
  generalization}.
\newblock \bibinfo{journal}{\emph{arXiv preprint arXiv:2004.05884}}
  (\bibinfo{year}{2020}).
\newblock


\bibitem[\protect\citeauthoryear{Wu, Tong, and Vorobeychik}{Wu
  et~al\mbox{.}}{2019}]%
        {wu2019defending}
\bibfield{author}{\bibinfo{person}{Tong Wu}, \bibinfo{person}{Liang Tong},
  {and} \bibinfo{person}{Yevgeniy Vorobeychik}.}
  \bibinfo{year}{2019}\natexlab{}.
\newblock \showarticletitle{Defending against physically realizable attacks on
  image classification}.
\newblock \bibinfo{journal}{\emph{arXiv preprint arXiv:1909.09552}}
  (\bibinfo{year}{2019}).
\newblock


\bibitem[\protect\citeauthoryear{Xiao, Li, Zhu, He, Liu, and Song}{Xiao
  et~al\mbox{.}}{2018}]%
        {xiao2018generating}
\bibfield{author}{\bibinfo{person}{Chaowei Xiao}, \bibinfo{person}{Bo Li},
  \bibinfo{person}{Jun-Yan Zhu}, \bibinfo{person}{Warren He},
  \bibinfo{person}{Mingyan Liu}, {and} \bibinfo{person}{Dawn Song}.}
  \bibinfo{year}{2018}\natexlab{}.
\newblock \showarticletitle{Generating adversarial examples with adversarial
  networks}.
\newblock \bibinfo{journal}{\emph{arXiv preprint arXiv:1801.02610}}
  (\bibinfo{year}{2018}).
\newblock


\bibitem[\protect\citeauthoryear{Xie, Wang, Zhang, Ren, and Yuille}{Xie
  et~al\mbox{.}}{2017}]%
        {xie2017mitigating}
\bibfield{author}{\bibinfo{person}{Cihang Xie}, \bibinfo{person}{Jianyu Wang},
  \bibinfo{person}{Zhishuai Zhang}, \bibinfo{person}{Zhou Ren}, {and}
  \bibinfo{person}{Alan Yuille}.} \bibinfo{year}{2017}\natexlab{}.
\newblock \showarticletitle{Mitigating adversarial effects through
  randomization}.
\newblock \bibinfo{journal}{\emph{arXiv preprint arXiv:1711.01991}}
  (\bibinfo{year}{2017}).
\newblock


\bibitem[\protect\citeauthoryear{Xie, Zhang, Zhou, Bai, Wang, Ren, and
  Yuille}{Xie et~al\mbox{.}}{2019}]%
        {xie2019improving}
\bibfield{author}{\bibinfo{person}{Cihang Xie}, \bibinfo{person}{Zhishuai
  Zhang}, \bibinfo{person}{Yuyin Zhou}, \bibinfo{person}{Song Bai},
  \bibinfo{person}{Jianyu Wang}, \bibinfo{person}{Zhou Ren}, {and}
  \bibinfo{person}{Alan~L Yuille}.} \bibinfo{year}{2019}\natexlab{}.
\newblock \showarticletitle{Improving transferability of adversarial examples
  with input diversity}. In \bibinfo{booktitle}{\emph{Proceedings of the
  IEEE/CVF Conference on Computer Vision and Pattern Recognition}}.
  \bibinfo{pages}{2730--2739}.
\newblock


\bibitem[\protect\citeauthoryear{Zhong, Squicciarini, Miller, and
  Caragea}{Zhong et~al\mbox{.}}{2017}]%
        {zhong2017group}
\bibfield{author}{\bibinfo{person}{Haoti Zhong}, \bibinfo{person}{Anna~Cinzia
  Squicciarini}, \bibinfo{person}{David~J Miller}, {and}
  \bibinfo{person}{Cornelia Caragea}.} \bibinfo{year}{2017}\natexlab{}.
\newblock \showarticletitle{A Group-Based Personalized Model for Image Privacy
  Classification and Labeling.}. In \bibinfo{booktitle}{\emph{IJCAI}},
  Vol.~\bibinfo{volume}{17}. \bibinfo{pages}{3952--3958}.
\newblock


\bibitem[\protect\citeauthoryear{Zhou, Liu, Han, Wang, Peng, and Gao}{Zhou
  et~al\mbox{.}}{2021}]%
        {zhou2021towards}
\bibfield{author}{\bibinfo{person}{Dawei Zhou}, \bibinfo{person}{Tongliang
  Liu}, \bibinfo{person}{Bo Han}, \bibinfo{person}{Nannan Wang},
  \bibinfo{person}{Chunlei Peng}, {and} \bibinfo{person}{Xinbo Gao}.}
  \bibinfo{year}{2021}\natexlab{}.
\newblock \showarticletitle{Towards Defending against Adversarial Examples via
  Attack-Invariant Features}.
\newblock \bibinfo{journal}{\emph{Proceedings of the 38 th International
  Conference on Machine Learning, (ICML)}} (\bibinfo{year}{2021}).
\newblock


\end{thebibliography}

\newpage
\appendix
\onecolumn

\section{Discussion on Reducing Perturbation Intensity}
\label{discussion on reducing}

\begin{figure}[ht]
	\centering
	\includegraphics[scale=0.35]{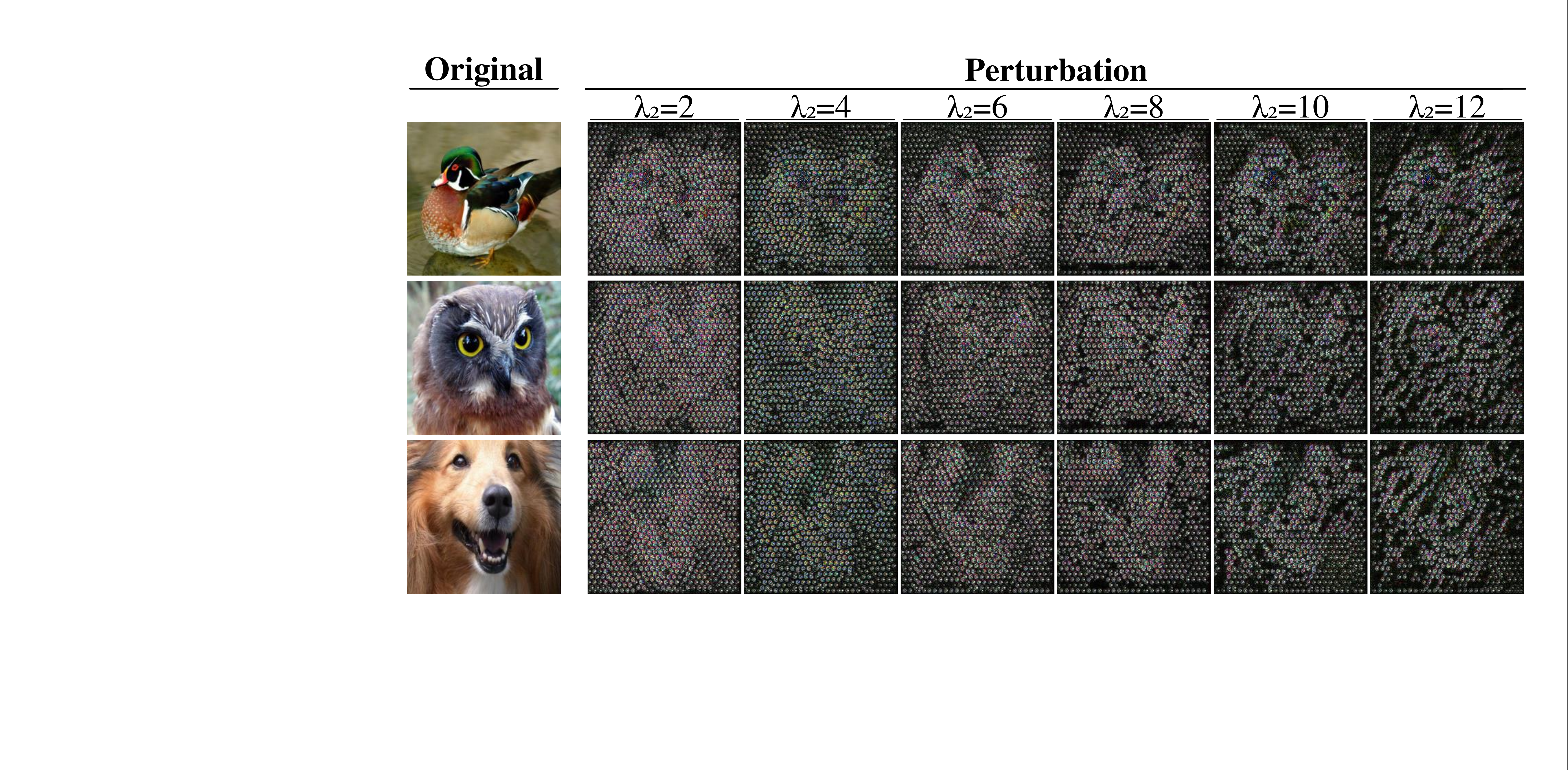}
	\caption{Adversarial perturbation generated by different weight $\lambda_2$ of $L_{G\_mse}$. With the increase of $\lambda_2$, the perturbation intensity becomes smaller.}
	\label{l2}
\end{figure}
\begin{figure}[ht]
	\centering
	\includegraphics[scale=0.6]{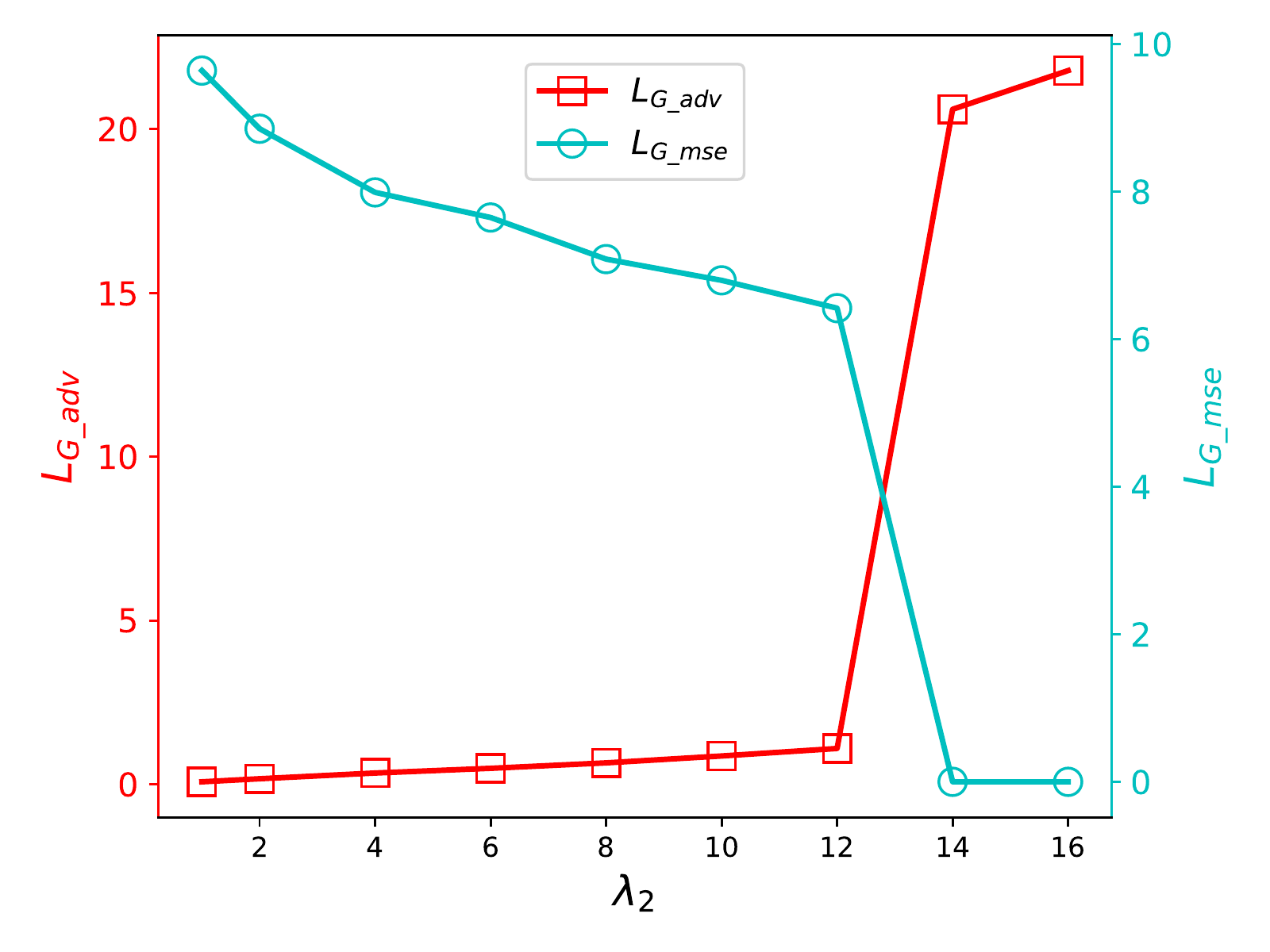}
	\caption{Line graph of $L_{G\_mse}$ and $L_{G\_adv}$. $L_{G\_mse}$ reflect the magnitude of perturbation and $L_{G\_adv}$ reflect the attack ability (lower is better). $\lambda_2$ is set to adjust the $L_{G\_mse}$.}
	\label{attack ability loss}
\end{figure}
As shown in Figure \ref{l2}, magnifying the weight $\lambda_2$ of $L_{G\_mse}$ does help to reduce the perturbation intensity. However, as shown in Figure \ref{attack ability loss}, the bigger  $\lambda_2$ brings smaller $L_{G\_mse}$ also brings the increase of $L_{G\_adv}$, which leads to the decrease of attack ability. Note that the generation-based method is different from the white-box attack. The generation-based method doesn't require parameters and structure of the target model after training. This difference makes the perturbation intensity important in ensuring the attack ability of the generation-based method. The smaller intensity is harder to optimize (e.g., when $\lambda_2=14$, the $L_{G\_adv}$ increases dramatically, which indicates the perturbation almost loses the adversarial property.)

\section{Proof}
\label{proof}
Here we provide a proof of $\Delta' <= \Delta$.

Given a image $x$ with $n$ pixels, supposing $r=[r_1,r_2,r_3,...,r_n] \sim R^n$ denote the perturbation adding to the each pixel of the image, $r'=[r'_1,r'_2,r'_3,...,r'_n] \sim R^n$ denote the perturbation $r$ pass the dimension reducer $DR$, and $r^*=[r^*_1,r^*_2,r^*_3,...,r^*_n] \sim R^n$ denote the recovered perturbation. Here we use $\Delta$ to represent the difference between $r$ and $r^*$ under $L_2$ norm, which can be described as
\begin{equation}
\begin{aligned}
\Delta &= ||r^*-r||_2\\
&=\sqrt{(r^*_1-r_1)^2+(r^*_2-r_2)^2+...+(r^*_n-r_n)^2}
\end{aligned}
\end{equation}
$\Delta'$ represent the difference between $r^*$ and $r'$ under $L_2$ norm, which can be described as
\begin{equation}
\begin{aligned}
\Delta'&= ||r^*-r'||_2 \\
&=\sqrt{(r^*_1-r'_1)^2+(r^*_2-r'_2)^2+...+(r^*_n-r'_n)^2} 
\end{aligned}
\end{equation} 

We take the average pooling with size $m$ (e.g., m=9 for the kernel size is 3$\times$3) as the down-sample and up-sample operation within the dimension reducer $DR$ for illustration, which means
\begin{equation}
r'_1=r'_2=...=r'_m = \overline{r}
\end{equation}
where
\begin{equation}
\overline{r}=\frac{1}{m}\sum_{i=1}^{m}r_i \quad(m \le n)
\end{equation}
Then, the difference between $ (\Delta'_m)^2$ and $(\Delta_m)^2$ can be calculated as
\begin{equation}
\begin{aligned}
(\Delta'_m)^2 - (\Delta_m)^2
&= \sum_{i=1}^{m}(r^*_i-r'_i)^2 - \sum_{i=1}^{m}(r^*_i-r_i)^2 \\
&=\sum_{i=1}^{m}(r^*_i-\overline{r})^2 - \sum_{i=1}^{m}(r^*_i-r_i)^2\\
&=\sum_{i=1}^{m}(r^*_i)^2-2\overline{r}\sum_{i=1}^{m}r^*_i+m\overline{r}^2
-\sum_{i=1}^{m}(r^*_i)^2+2\sum_{i=1}^{m}r^*_ir_i-\sum_{i=1}^{m}(r_i)^2 \\
&=-2m\overline{r}\overline{r^*}+m\overline{r}^2+2m\overline{r}\overline{r^*}-m\overline{r^2}\\
&=m\overline{r}^2-m\overline{r^2}
\end{aligned}
\label{delta1}
\end{equation}
Meanwhile,
\begin{equation}
\begin{aligned}
\sum_{i=1}^{m}(r_i-\overline{r})^2 &\ge 0 \\
\sum_{i=1}^{m}(r_i^2+\overline{r}^2-2r_i^2\overline{r}) &\ge 0 \\
\sum_{i=1}^{m}r_i^2+m\overline{r}^2-2m\overline{r}^2 &\ge 0 \\
\sum_{i=1}^{m}r_i^2-m\overline{r}^2 &\ge 0 \\
m\overline{r^2} &\ge m\overline{r}^2
\end{aligned}
\label{delta avg1}
\end{equation}
With the combination of Eq.\ref{delta1} and Eq. \ref{delta avg1}, we can obtain $(\Delta'_m)^2 <= (\Delta_m)^2$. Because $\Delta'_m>=0$ and $\Delta_m>=0$, we can obtain $\Delta'_m <= \Delta_m$ within each block. Thus, for the whole image, we can obtain $\Delta' <= \Delta$.

\newpage

\section{The Recoverability of Other Attacks}
\label{other recoverability}
\begin{table*}[h]
	\caption{Recoverability comparison between state-of-the-art defense against C\&W attack and our RGAN on MNIST and Caltech-256. Our method achieves better recoverability across different datasets and different network architectures.}
	\setlength{\tabcolsep}{1.2mm}
	\begin{tabular}{ccccccccccccc}
		\toprule
		& \multicolumn{3}{c}{MNIST}                        & \multicolumn{9}{c}{Caltech-256}\\ 
		\cmidrule(lr){2-4} \cmidrule(lr){5-13} 
		& \multicolumn{3}{c}{LeNet-5}                      & \multicolumn{3}{c}{DenseNet-121}                  & \multicolumn{3}{c}{ResNet-50}                     & \multicolumn{3}{c}{MoblieNetV3}                   \\ \cmidrule(lr){2-4} \cmidrule(lr){5-7} \cmidrule(lr){8-10} \cmidrule(lr){11-13}
		& $L_2$            & PSNR(dB)            & CER(\%)      & $L_2$            & PSNR(dB)             & CER(\%)         & $L_2$            & PSNR(dB)             & CER(\%)         & $L_2$            & PSNR(dB)             & CER(\%)         \\ \midrule
		Xie $et$   $al.$\cite{xie2017mitigating}         & 7.66          & 11.59            & 39.14         & 88.75         & 13.59            & 21.75          & 88.82         & 13.57            & 21.67          & 89.24         & 13.52            & 29.57          \\
		Guo $et$   $al.$\cite{guo2017countering}         & 2.95          & 19.67            & 19.91         & 11.26         & 30.63            & 22.76          & 12.63         & 30.23            & 24.43          & 11.33         & 30.56            & 30.03          \\
		Prakash $et$   $al.$\cite{prakash2018deflecting} & 4.40          & 16.20            & 36.18         & 8.84          & 32.92            & 22.37          & 10.58         & 32.39            & 22.76          & 8.93          & 32.89            & 28.98          \\
		Das $et$   $al.$\cite{das2018shield}             & 1.54          & 25.48            & 1.43          & 7.70          & 34.60            & 17.82          & 9.23          & 34.05            & 20.03          & 7.79          & 34.56            & 23.15          \\
		Mustafa $et$   $al.$\cite{mustafa2019image}      & 1.91          & 23.70            & 2.77          & 11.04         & 31.03            & 25.44          & 12.23         & 30.67            & 24.12          & 11.10         & 31.02            & 29.49          \\
		Jin  $et$ $al.$\cite{jin2019ape}                 & 1.63          & 24.82            & 1.45          & 25.28         & 23.86            & 23.65          & 19.74         & 26.43            & 25.68          & 25.78         & 23.82            & 57.58          \\
		Zhou  $et$ $al.$\cite{zhou2021towards}           & 1.09          & 28.38            & 1.30          & 39.39         & 20.34            & 93.85          & 39.51         & 20.34            & 93.42          & 38.60         & 20.57            & 92.37          \\
		RGAN(our)                                                         & \textbf{0.50} & \textbf{35.24}   & \textbf{1.16} & \textbf{1.29} & \textbf{48.09}   & \textbf{17.19} & \textbf{2.44} & \textbf{43.73}   & \textbf{16.92} & \textbf{1.84} & \textbf{45.97}   & \textbf{21.36} \\ \midrule
		Clean                                                                & 0.00          & $+\infty$ & 1.11          & 0.00          & $+\infty$ & 17.00          & 0.00          & $+\infty$ & 16.34          & 0.00          & $+\infty$ & 20.85          \\ 
		\bottomrule
	\end{tabular}
	\label{defense attack cw}
\end{table*}

\begin{table*}[ht]
	\caption{Recoverability comparison between state-of-the-art defense against PGD attack and our RGAN on MNIST and Caltech-256. Our method achieves better recoverability across different datasets and different network architectures.}
	\setlength{\tabcolsep}{1.2mm}
	\begin{tabular}{ccccccccccccc}
		\toprule
		& \multicolumn{3}{c}{MNIST}                        & \multicolumn{9}{c}{Caltech-256}   \\ 
		\cmidrule(lr){2-4} \cmidrule(lr){5-13} 
		& \multicolumn{3}{c}{LeNet-5}                      & \multicolumn{3}{c}{DenseNet-121}                  & \multicolumn{3}{c}{ResNet-50}                     & \multicolumn{3}{c}{MoblieNetV3}                   \\ \cmidrule(lr){2-4} \cmidrule(lr){5-7} \cmidrule(lr){8-10} \cmidrule(lr){11-13}
		& $L_2$            & PSNR(dB)            & CER(\%)      & $L_2$            & PSNR(dB)             & CER(\%)         & $L_2$            & PSNR(dB)             & CER(\%)         & $L_2$            & PSNR(dB)             & CER(\%)         \\ \midrule
		Xie $et$ $al.$\cite{xie2017mitigating}         & 8.09          & 10.89            & 97.87         & 89.78         & 13.44            & 97.35          & 88.53         & 13.58            & 92.52          & 89.34         & 13.48            & 96.10          \\
		Guo $et$   $al.$\cite{guo2017countering}         & 5.49          & 14.18            & 100.00        & 12.59         & 30.16            & 95.99          & 12.60         & 30.15            & 94.04          & 12.64         & 30.13            & 98.32          \\
		Prakash $et$   $al.$\cite{prakash2018deflecting} & 6.68          & 12.49            & 100.00        & 9.56          & 32.17            & 83.89          & 9.55          & 32.18            & 78.91          & 9.63          & 32.11            & 84.98          \\
		Das $et$   $al.$\cite{das2018shield}             & 5.65          & 13.90            & 100.00        & 9.73          & 32.23            & 95.56          & 9.74          & 32.22            & 89.22          & 10.05         & 31.92            & 99.10          \\
		Mustafa $et$   $al.$\cite{mustafa2019image}      & 5.39          & 14.30            & 100.00        & 11.46         & 30.69            & 78.13          & 11.45         & 30.70            & 73.89          & 11.44         & 30.70            & 72.76          \\
		Jin  $et$ $al.$\cite{jin2019ape}                 & 2.33          & 21.74            & 4.80          & 26.89         & 23.63            & 50.93          & 58.97         & 16.71            & 82.21          & 25.78         & 23.82            & 57.58          \\
		Zhou  $et$ $al.$\cite{zhou2021towards}           & 1.48          & 25.71            & 2.25          & 39.55         & 20.29            & 94.04          & 39.38         & 20.37            & 93.65          & 38.57         & 20.57            & 92.41          \\
		RGAN(our)                                                         & \textbf{0.50} & \textbf{35.24}   & \textbf{1.16} & \textbf{1.29} & \textbf{48.09}   & \textbf{17.19} & \textbf{2.44} & \textbf{43.73}   & \textbf{16.92} & \textbf{1.84} & \textbf{45.97}   & \textbf{21.36} \\ \midrule
		Clean                                                                & 0.00          & $+\infty$ & 1.11          & 0.00          & $+\infty$ & 17.00          & 0.00          & $+\infty$ & 16.34          & 0.00          & $+\infty$ & 20.85          \\ \bottomrule
	\end{tabular}
	\label{defense attack pgd}
\end{table*}

\newpage
\section{Additional Images}
\label{addition images}
\begin{figure}[h]
	\centering
	\includegraphics[scale=0.4]{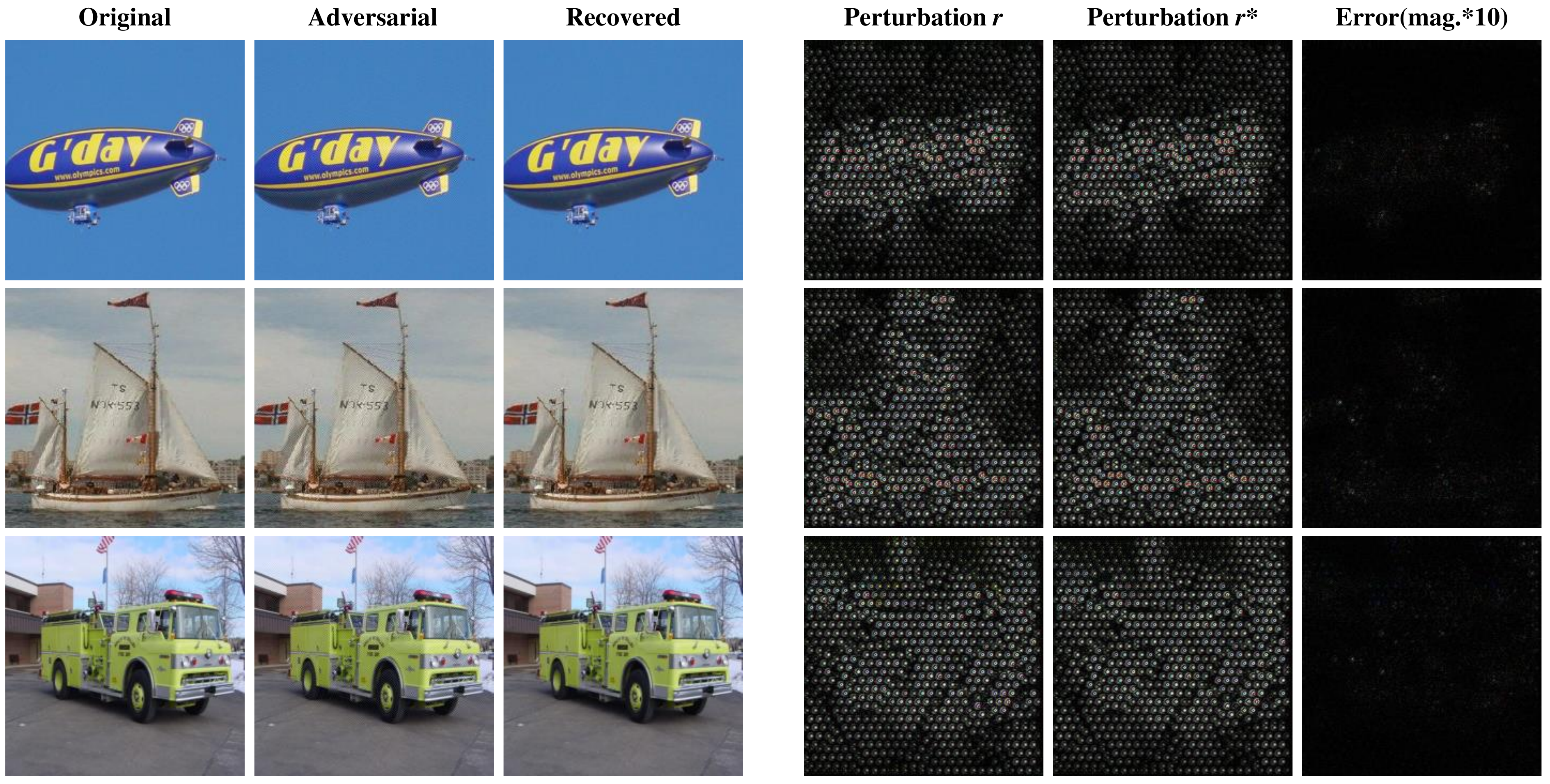}
	\caption{More examples similar to Figure \ref{demo1}. From the left: original example, adversarial examples, recovered examples, adversarial perturbation $r$, recovered perturbation $r*$, and error between the original and recovered examples.}
	\label{demo2}
\end{figure}

%%
%% The acknowledgments section is defined using the "acks" environment
%% (and NOT an unnumbered section). This ensures the proper
%% identification of the section in the article metadata, and the
%% consistent spelling of the heading.

\end{document}